\documentclass[runningheads]{llncs}
\usepackage{graphicx}
\usepackage{comment}
\usepackage{amsmath,amssymb} 
\usepackage{color}
\usepackage{booktabs}
\usepackage[colorlinks,linkcolor=blue]{hyperref}
\usepackage{multirow}
\usepackage{threeparttable}
\usepackage{subfigure}
\usepackage{overpic}
\newcommand{\Paragraph}[1]{\vspace{2mm} \noindent \textbf{#1} \hspace{0mm}}

\usepackage[width=122mm,left=12mm,paperwidth=146mm,height=193mm,top=12mm,paperheight=217mm]{geometry}

\newcommand\blfootnote[1]{%
	\begingroup
	\renewcommand\thefootnote{}\footnote{#1}%
	\addtocounter{footnote}{-1}%
	\endgroup
}

\begin{document}
\pagestyle{headings}
\mainmatter
\def\ECCVSubNumber{2778}  

\title{ASFD: Automatic and Scalable Face Detector} 

\titlerunning{ASFD: Automatic and Scalable Face Detector}
%

\author{Bin Zhang${}^\ast$\inst{1,2} \and Jian Li${}^\ast$\inst{1} \and
Yabiao Wang\inst{1} \and Ying Tai\inst{1} \and Chengjie Wang\inst{1} \and Jilin Li\inst{1} \and Feiyue Huang\inst{1} \and Yili Xia\inst{2} \and Wenjiang Pei\inst{2} \and Rongrong Ji\inst{3}}
\authorrunning{B. Zhang, J. Li, and et al.}
%
\institute{Youtu Lab, Tencent \and
School of Information Science and Engineering, Southeast University, China \and
Artificial Intelligence Department, Xiamen University, China\\
\email{\{z-bingo, yili\_xia, wjpei\}@seu.edu.cn, \{swordli, caseywang, yingtai, jasoncjwang, jerolinli, garyhuang\}@tencent.com, rrji@xmu.seu.cn}}
\maketitle

\blfootnote{\scriptsize ${}^\ast$ These authors contributed equally. This work was doen when Bin Zhang was an intern as Tencent Youtu Lab.}

\vspace{-30pt}
\begin{abstract}
In this paper, we propose a novel Automatic and Scalable Face Detector (ASFD), which is based on a combination of neural architecture search techniques as well as a new loss design. 
First, we propose an automatic feature enhance module named Auto-FEM by improved differential architecture search, which allows efficient multi-scale feature fusion and context enhancement. 
Second, we use Distance-based Regression and Margin-based Classification (DRMC) multi-task loss to predict accurate bounding boxes and learn highly discriminative deep features. 
Third, we adopt compound scaling methods and uniformly scale the backbone, feature modules, and head networks to develop a family of ASFD, which are consistently more efficient than the state-of-the-art face detectors. 
Extensive experiments conducted on popular benchmarks, \textit{e.g.} WIDER FACE and FDDB, demonstrate that our ASFD-D$6$ outperforms the prior strong competitors, and our lightweight ASFD-D$0$ runs at more than $120$ FPS with Mobilenet for VGA-resolution images.

\vspace{-6pt}
\end{abstract}

\vspace{-26pt}
\section{Introduction}
\vspace{-4pt}
Face detection is the prerequisite step of facial image analysis for various applications, such as face alignment~\cite{tai2019towards}, attribute~\cite{zhang2018joint,pan2018mean}, recognition~\cite{yang2016nuclear,Huang2020curricularface} and verification~\cite{deng2019arcface,wang2018cosface}. 
In the past few years, tremendous progress has been made on designing the model architecture of deep Convolutional Neural Networks (CNNs)~\cite{he2016deep} for face detection. 
However, it remains a challenge to accurately detect faces with a high degree of variability in scale, pose, occlusion, expression, appearance, and illumination. 
In addition, the large model sizes and expensive computation costs make these detectors difficult to be deployed in many real-world applications where memory and latency are highly constrained.

\begin{figure}[!t]
    \centering
    \subfigure[]{
    \includegraphics[height=0.25\linewidth]{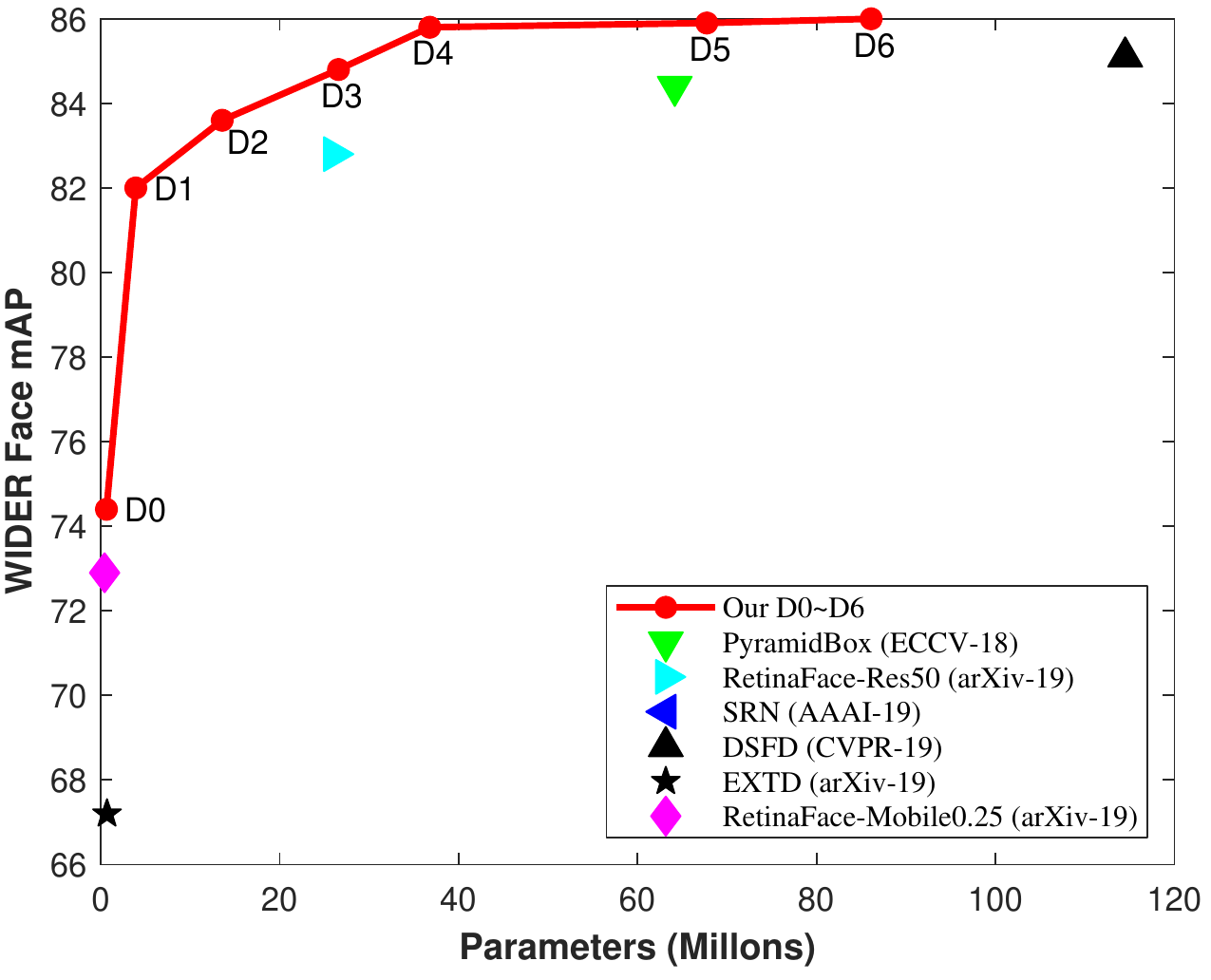}
    }
    \subfigure[]{
    \includegraphics[height=0.25\linewidth]{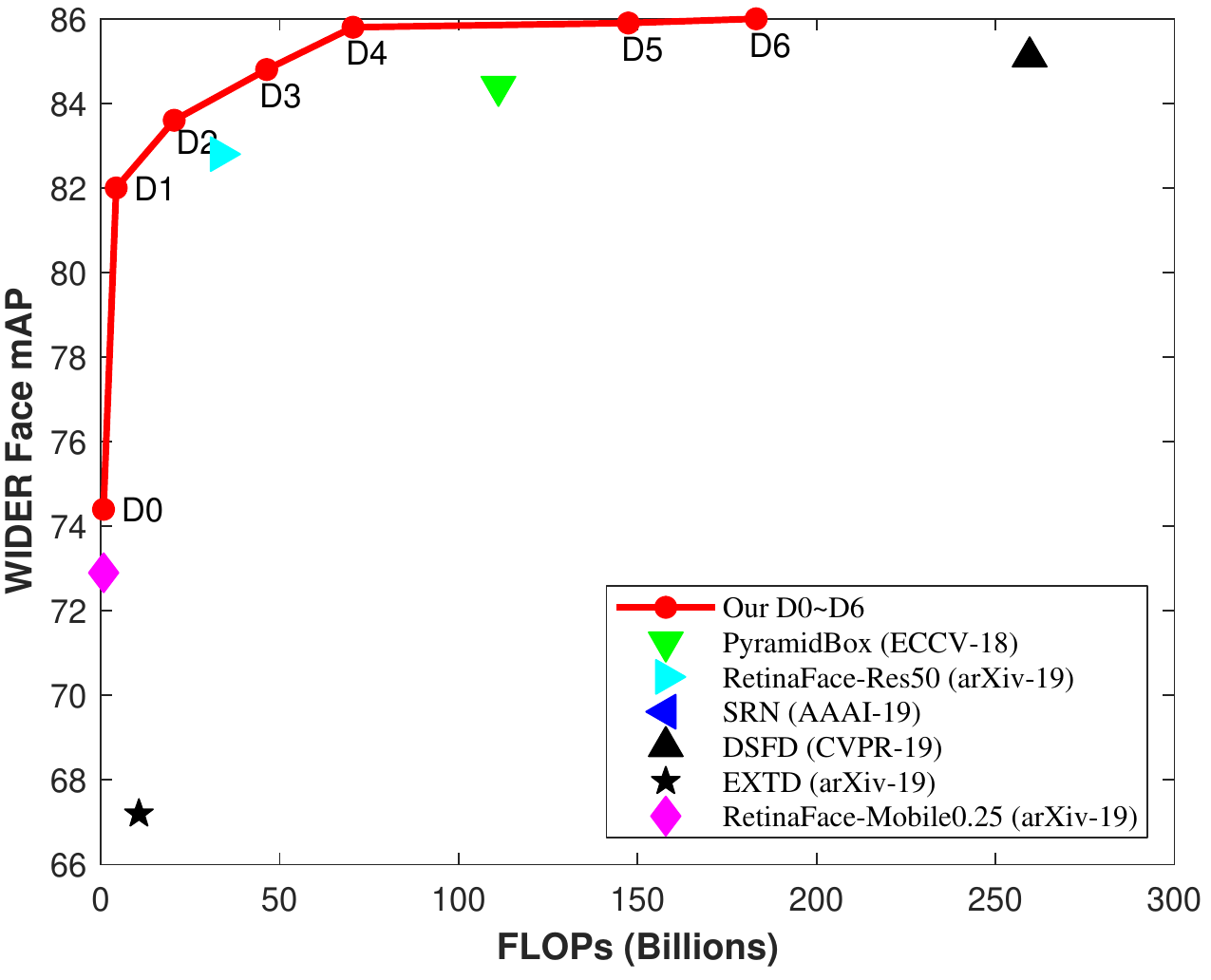}
    }
    \subfigure[]{
    \includegraphics[height=0.25\linewidth]{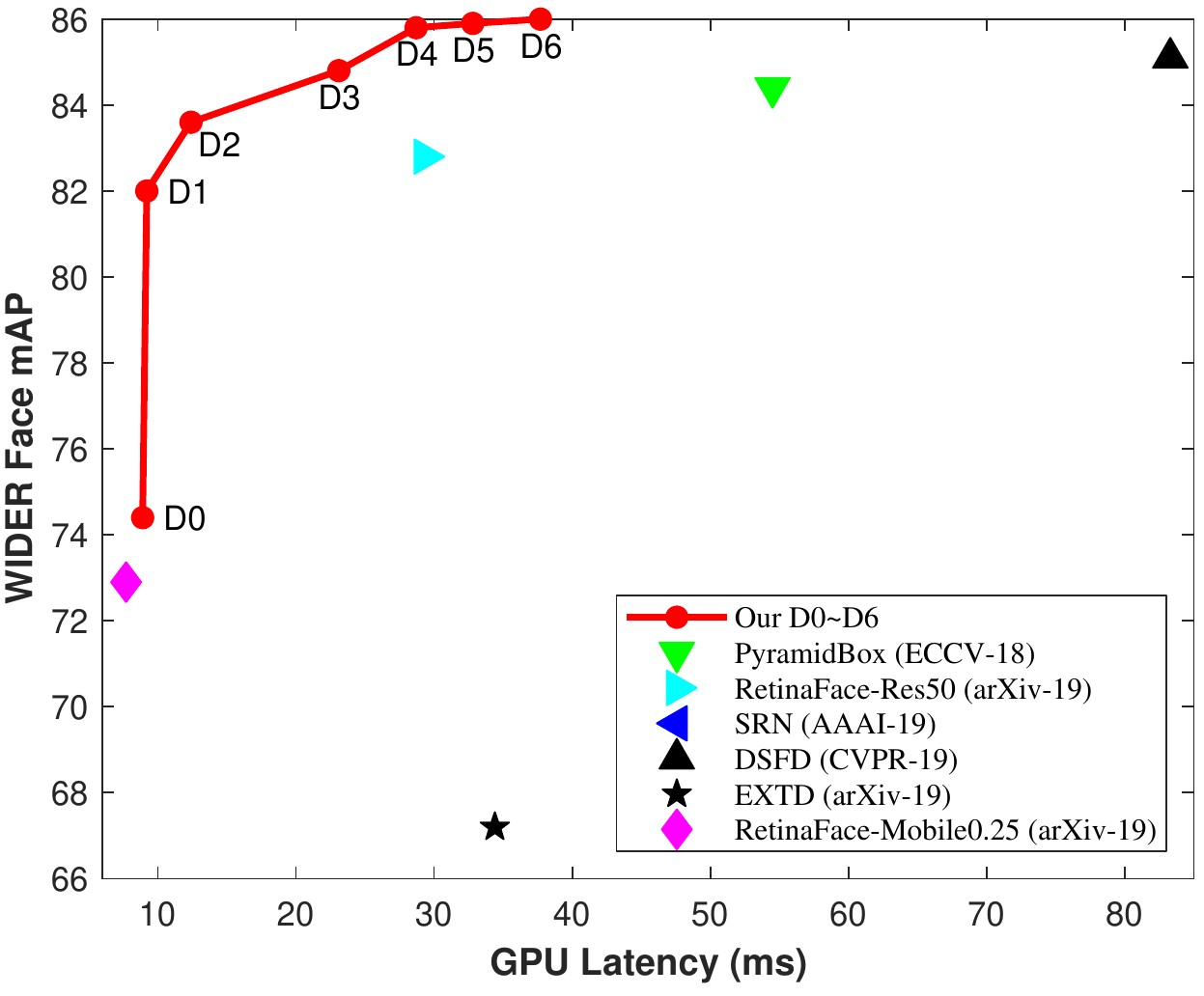}
    }
    \vspace{-3mm}
    \caption{\small \textbf{Illustration of the mean Average Precision (mAP) regarding the number of parameters (a), FLOPs (b) and GPU latency (c)} evaluated with single-model single-scale on the validation subset of WIDER FACE dataset, where mAP is equivalent to the AP of Hard set. Our ASFD D$0$-D$6$ outperforms the prior detectors with respect to parameter numbers, FLOPs, and latency.}
    \vspace{-4mm}
    \label{fig:params_vs_perf}
\end{figure}

There have been many works aiming to develop face detector architectures, mainly composed of one-stage~\cite{tang2018pyramidbox,chi2019selective,li2019dsfd,deng2019retinaface,zhang2019refineface} and two-stage~\cite{wang2017face,wang2017detecting,zhang2018face} face detectors. 
Among them, the one-stage is the domain- and anchor-based face detection approach, which tiles regular and dense anchors with various scales and aspect ratios over all locations of several multi-scale feature maps. 
Generally, there are four key-parts in this framework, including backbone, feature module, head network, and multi-task loss. 
Feature module uses Feature Pyramid Network (FPN)~\cite{lin2017feature,li2017object} to aggregate hierarchical feature maps between high- and low-level features of backbone, and the module for refining the receptive field~\cite{liu2018receptive,li2019dsfd,zhang2019refineface}, such as Receptive Field Block (RFB), is also introduced to provide rich contextual information for hard faces. 
Moreover, multi-task loss is composed of the binary classification and bounding box regression, in which the former classifies the predefined anchors into face and background, and the latter regresses those detected faces to accurate locations. 
Despite the progress achieved by above methods, there are still some problems existed in three aspects:

\textbf{Feature Module.} Although FPN~\cite{lin2017feature} and RFB~\cite{liu2018receptive} are simple and effective for general object detection, they may be suboptimal for face detection and 
many recent works~\cite{li2019dsfd,zhang2019refineface} propose various cross-scale connections or operations to combine features to generate the better representations. 
However, the challenge still exists in the huge design space for feature module. 
In addition, these methods all adopt the same feature modules for different feature maps from the backbone, which ignore the importance and contributions of different input features. 

\textbf{Multi-task Loss.}
The conventional multi-task loss in object detection includes a regression loss and a classification loss~\cite{girshick2015fast,ren2015faster,liu2016ssd,lin2017focal}. 
Smooth-L$_1$ loss for the bounding box regression is commonly used in current face detectors~\cite{li2019dsfd,zhang2019refineface}, which however achieves slow convergence and inaccurate regression for its sensitivity to variant scales.
As for the classification, standard binary softmax loss in DSFD~\cite{li2019dsfd} usually lacks the power of discrimination, 
and RefineFace~\cite{zhang2019refineface} adopts sigmoid focal loss for better distinguishing faces from the background, which relies on predefined hyper-parameters and is extremely time-consuming. 

\textbf{Efficiency and Accuracy.}
Both DSFD and RefineFace rely on the big backbone networks, deep detection head networks and large input image sizes for high accuracy, while FaceBox~\cite{zhang2017faceboxes} is a lightweight face detector with fewer layers to achieve better efficiency by sacrificing accuracy. 
The above methods can not balance the efficiency and accuracy in a wide spectrum of resource constraints from mobile devices to data centers in real-world applications. 
An appropriate selection of network width and depth usually require tedious manual tuning. 

To address these issues, we propose a novel Automatic and Scalable Face Detector (ASFD) to deliver the next generation of efficient face detector with high accuracy. 
Specifically, we first introduce an Automatic Feature Enhance Module (Auto-FEM) via improved differential architecture search to exploit feature module for efficient and effetive multi-scale feature fusion and context enhancement.
Second, inspired by distance Intersection over Union (IoU) loss~\cite{zheng2019distance} and large margin cosine loss~\cite{wang2018cosface}, we propose a Distance-based Regression and Margin-based Classification (DRMC) multi-task loss for accurate bounding boxes and highly discriminative deep features.
Finally, motivated by scalable model design described in EfficientNet~\cite{tan2019efficientnet} and EfficientDet~\cite{tan2019efficientdet}, We adopt compound scaling methods and uniformly scale the backbone, feature module and head networks to develop a family of our ASFD, which consistently outperforms the prior competitors in terms of parameter numbers, FLOPs and latency, as shown in Fig.~\ref{fig:params_vs_perf}, achieving the better trade-off between efficiency and accuracy.

In summary, the main contributions of this paper include:

$\bullet$ Automatic Feature Enhance Module via improved differential architecture search for efficient multi-scale feature fusion and context enhancement.

$\bullet$ Distance-based regression and margin-based classification multi-task loss for accurate bounding boxes and highly discriminative deep features.

$\bullet$ A new family of face detectors 
achieved by compound scaling methods on backbone, feature module, head network and resolution.

$\bullet$ Comprehensive experiments conducted on popular benchmarks, \textit{e.g.} WIDER FACE and FDDB, to demonstrate the efficiency and accuracy of our ASFD compared with state-of-the-art methods.

\vspace{-10pt}
\section{Related Work}
\vspace{-10pt}
\Paragraph{Face detection.}
Traditional face detection methods mainly rely on hand-crafted features, such as Haar-like features~\cite{viola2004robust}, control point set~\cite{abramson2005yef} and edge orientation histograms~\cite{levi2004learning}. 
With the development of deep learning, Overfeat~\cite{sermanet2013overfeat}, Cascade-CNN~\cite{li2015cascadecnn}, MTCNN~\cite{zhang2016mtcnn} adopt CNN to classify sliding window, which is not end-to-end and inefficient. 
Current state-of-the-art face detection methods have inherited some achievements from generic object detection~\cite{ren2015faster,liu2016ssd,lin2017focal,zhang2018refinedet} approaches. 
More recently, DSFD~\cite{li2019dsfd} and Refineface~\cite{zhang2019refineface} propose pseudo two-stage structure based on single-shot framework to make face detector more effective and accurate. 
There are mainly two differences between the previous face detectors and our ASFD: 
($1$) Automatic feature module is obtained by improved NAS method instead of hand-designed.
($2$) The margin-based loss and distance-based loss are employed together for the power of discrimination.

\Paragraph{Neural Architecture Search.}
Neural architecture search (NAS) has attracted increasing research interests. NASNet~\cite{zoph2018learning} uses Reinforcement Learning (RL) with a controller Recurrent Neural Network (RNN) to search neural architectures sequentially. 
To save computational resources, Differential Architecture Search (DARTS)~\cite{liu2018darts} is based on continuous relaxation of a supernet and propose gradient-based search. Partially-Connected DARTS (PC-DARTS)~\cite{xu2019pcdarts} samples a small part of supernet to reduce the redundancy in network space, Based on above NAS works on image classification, some recent works attempt to develop NAS to generic object detection. DetNAS~\cite{chen2019detnas} tries to search better backbones for object detection, while NAS-FPN~\cite{ghiasi2019fpn} targets on searching for an FPN alternative based on RNN and RL, which is time-consuming. NAS-FCOS~\cite{wang2019fcos} aims to efficiently search for the FPN as well as the prediction head based on anchor-free one-stage framework. 
Different from DARTS or PC-DARTS, we introduce an improved NAS which only samples the path with the highest weight for each node during the forward pass of the searching phase to further reduce the memory cost.
To our best knowledge, ASFD is the first work to report the success of applying differential architecture search in face detection community. 

\Paragraph{Model Scaling.}
There are several approaches to scale a network, for instance,
ResNet~\cite{he2016deep} can be scaled down (\textit{e.g.}, ResNet-$18$) or up (\textit{e.g.}, ResNet-$200$) by adjusting network depth. 
Recently, EfficientNet~\cite{tan2019efficientnet} demonstrates remarkable model efficiency for image classification by jointly scaling up network width, depth, and resolution.
For object detection, EfficientDet~\cite{tan2019efficientdet} proposes a compound scaling method that uniformly scales the resolution, depth and width for all backbone, feature network, and box/class prediction networks at the same time. 
Inspired by the above model scaling methods, we develop a new family of face detectors, \textit{i.e.} ASFD D$0$-D$6$, to optimize both accuracy and efficiency. 

\vspace{-8pt}
\section{Our approach}
\vspace{-4pt}
We firstly introduce the pipeline of our proposed framework in Sec.~\ref{section:3.1}, then describe our automatic feature enhance module in Sec.~\ref{section:3.2}, distance-based regression and margin-based classification loss in Sec.~\ref{section:3.3}. At last, based on the improved model scaling, we develop a new family of face detectors in Sec.~\ref{section:3.4}.

\begin{figure}[!t]
\centering
\includegraphics[trim={0 0 0 0mm},clip,width=0.95\linewidth]{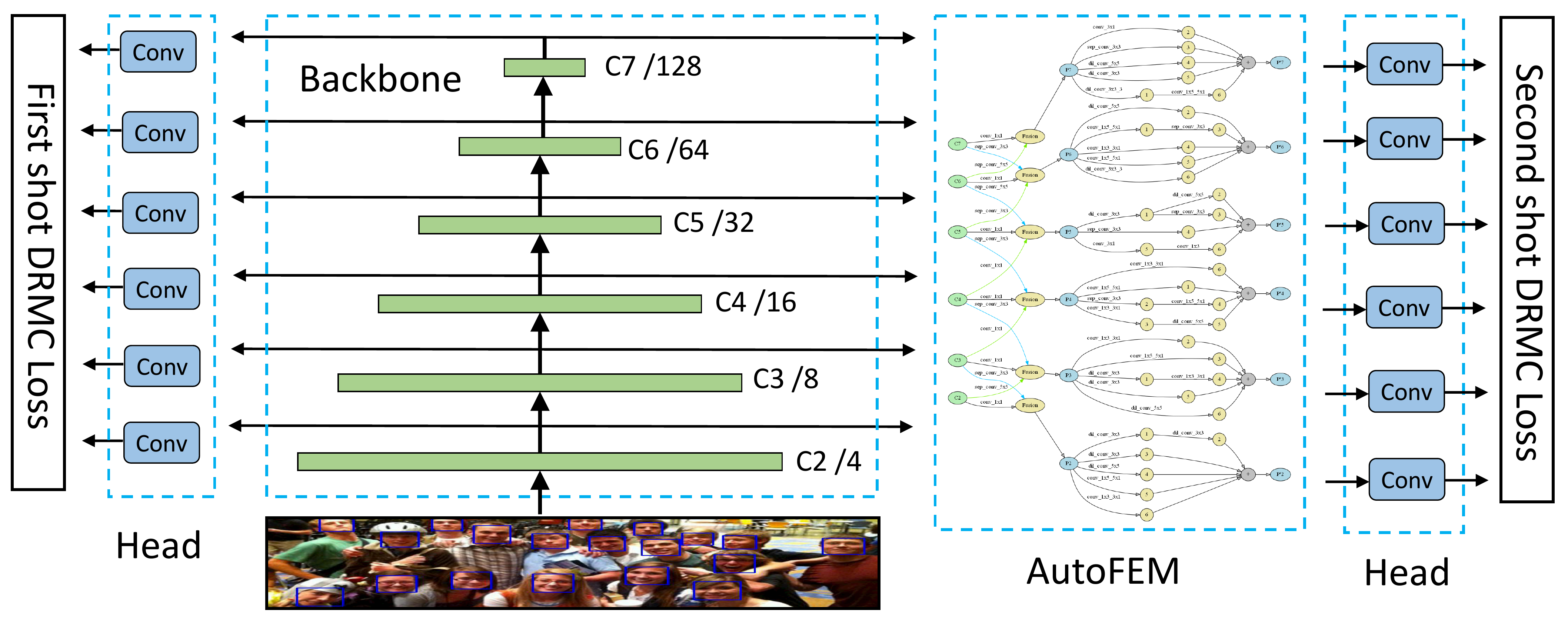}
\vspace{-3mm}
\caption{\small \textbf{Illustration on the framework of ASFD}. 
We propose an \textbf{AutoFEM} on right lateral of a feedforward backbone to generate the enhanced features. 
The original and enhanced features adopt our proposed \textbf{DRMC} loss.}
\vspace{-3mm}
\label{fig:framework}
\vspace{-3mm}
\end{figure}

\vspace{-6pt}
\subsection{Pipeline}\label{section:3.1}
\vspace{-2pt}
Fig.~\ref{fig:framework} illustrates the overall framework of ASFD, which follows the paradigm of DSFD~\cite{li2019dsfd} using the dual shot structure. 
The ImageNet-pretrained backbone generates six pyramidal feature maps $\{\mathbf{C}_2, \mathbf{C}_3, \mathbf{C}_4, \mathbf{C}_5, \mathbf{C}_6, \mathbf{C}_7\}$, whose stride varies from $4$ to $128$. 
Our proposed AutoFEM transfers these original feature maps into six enhanced feature maps. 
Both regression and classification head networks consist of several convolutions and map the original and enhanced features to produce class and bounding box. In particular, the two shots share the same head network and adopt the proposed DRMC loss for optimizing. 

\vspace{8pt}
\textbf{Details of our method will be released later, stay tuned please.}

\vspace{-4pt}
\section{Experiments}
\subsection{Implementation Details}
During training, we use ImageNet pretrained models to initialize the backbone. 
SGD optimizer is applied to fine-tune the models with $0.9$ momentum, $0.0005$ weight decay and batch size $48$ on four Nvidia Tesla V$100$ ($32$GB) GPUs. 
The learning rate is linearly risen from $10^{-6}$ to $0.015$ at the first $500$ iterations using the warmup strategy, then divided by $10$ at $25$ and $40$ epochs and ending at $50$ epochs. 
For inference, non-maximum suppression is applied with Jaccard overlap of $0.3$ to produce the top $750$ high confident faces from $5000$ high confident detections. 
All models are only trained on the training set of WIDER FACE.

In the search scenario, ResNet$50$ is selected as the backbone of our supernet, the channels of both AutoFEM-FPN and AutoFEM-CPM are set to $256$, and each AutoFEM-CPM consists of $6$ intermediate nodes. 
For efficiency, only $1/4$ features are sampled on each edge following the setting of PC-DARTS. 
We use Adam with learning rate $0.01$ and weight decay $0.0005$ to optimize the architecture parameters after $20$ epochs, and total searching epoch number is $50$.

\begin{table}[!tbp]
\small
\begin{center}
\caption{\small \textbf{Comparison of Average Precision (AP) among AutoFEM and state-of-the-art structures} on validation set of WIDER FACE. Multi-scale results ensemble is adopted during test-time.}
\resizebox{.95\columnwidth}{!}{
\smallskip\begin{tabular}{c|cccccccccccccc}
\toprule[2pt]
Feature module & \multicolumn{14}{c}{Baseline and Contributions}  \\
\midrule[1pt]
FEM-FPN~\cite{li2019dsfd}    &  & \checkmark &  &  &  &  &  &  &\checkmark & \checkmark & \checkmark\\
BiFPN~\cite{tan2019efficientdet}      &  &  & \checkmark &  &  &  &  &  & \\
PAN~\cite{liu2018path}      &  &  &  & \checkmark &  &  &  &  & \\
AutoFEM-FPN &  &  &  &  & \checkmark &  &  &  &  &  &  &\checkmark & \checkmark & \checkmark\\ \hline
 FEM-CPM~\cite{li2019dsfd}  &  &  &  &  &  & \checkmark &  &  & \checkmark &  &  & \checkmark \\
RFE~\cite{zhang2019refineface}        &  &  &  &  &  &  & \checkmark &  &  & \checkmark &  &  & \checkmark\\
AutoFEM-CPM &  &  &  &  &  &  &  &\checkmark &  &  & \checkmark &  &  & \checkmark\\
\midrule[1pt]\smallskip
Easy   & 0.947 & 0.954 & 0.954 & 0.953 & 0.956 & 0.950 & 0.951 & 0.948 & 0.954 & 0.954 & 0.952 & 0.955 & 0.956 & \textbf{0.958} \\\smallskip
Medium & 0.932 & 0.944 & 0.945 & 0.945 & 0.947 & 0.933 & 0.934 & 0.933 & 0.944 & 0.945 & 0.943 & 0.945 & 0.947 & \textbf{0.949}  \\\smallskip
Hard   & 0.822 & 0.881 & 0.874 & 0.883 & 0.884 & 0.827 & 0.830 & 0.834 & 0.882 & 0.882 & 0.881 & 0.886 & 0.886 & \textbf{0.887} \\
\toprule[2pt]
\end{tabular}
}
\vspace{-6mm}
\label{table:autofem}
\end{center}
\end{table}

\vspace{-4pt}
\subsection{Analysis on ASFD}\label{sec:simulations}
\vspace{-2pt}
\subsubsection{AutoFEM.} The architectures of AutoFEM-FPN and AutoFEM-CPM are searched on the basis of light DSFD~\cite{li2019dsfd} respectively, which uses ResNet$50$ as backbone and has $2$-layer head modules. 
The proposed AutoFEM is obtained by cascading these two modules together, an example of AutoFEM is presented in Fig.~\ref{fig:autofem}, which is adopted in our ASFD. As for AutoFEM-FPN, each output level fuses the features from its neighbor levels with varied convolution and its same levels with $1\times 1$ convolution, suggesting the importance of information from bottom and up layers, and different context prediction modules are obtained for 6 detection layers, in which the low-level CPM have larger receptive fields for capturing more context to improve performance of hard faces.

To demonstrate the effectiveness of our searched AutoFEM in ASFD, experiments are conducted to compare our AutoFEM-FPN and AutoFEM-CPM with other state-of-the-art structures. 
The DSFD-based detector with backbone of ResNet$50$ without FEM module is employed as the baseline, and all experimental results of applying feature pyramid network and context prediction module to feature module are shown in Table~\ref{table:autofem}, which indicates our AutoFEM improves the detection performance. 
It is obvious that after using the AutoFEM-FPN, the AP scores of the baseline detector are improved from $94.7\%$, $93.2\%$, $82.2\%$ to $95.6\%$, $94.7\%$, $88.4\%$ on the Easy, Medium and Hard subsets, respectively, which surpasses other structures like FEM-FPN~\cite{li2019dsfd}, BiFPN~\cite{tan2019efficientdet} and PAN~\cite{liu2018path}, and the performance is further improved to $95.8\%$, $94.9\%$, $88.7\%$ by cascading AutoFEM-FPN and AutoFEM-CPM together.

\begin{table}[!t]
\small
\begin{center}
\caption{\small \textbf{Comparison of Average Precision (AP) of PC-DARTS and our improved method for searching AutoFEM-CPM} evaluated on validation set of WIDER FACE. Multi-scale results ensemble is adopted during test-time.}
\resizebox{0.75\columnwidth}{!}{
\begin{threeparttable}
\begin{tabular}{@{}c|ccc|ccc|ccc@{}}
\toprule[2pt]
\multirow{2}{*}{Method} & \multicolumn{3}{c|}{4 inter nodes} & \multicolumn{3}{c|}{6 inter nodes} & \multicolumn{3}{c}{8 inter nodes} \\ \cmidrule(l){2-10} 
    & \multicolumn{1}{c|}{Easy} & \multicolumn{1}{c|}{Medium} & \multicolumn{1}{c|}{Hard} & \multicolumn{1}{c|}{Easy} & \multicolumn{1}{c|}{Medium} & \multicolumn{1}{c|}{Hard} & \multicolumn{1}{c|}{Easy} & \multicolumn{1}{c|}{Medium} & \multicolumn{1}{c}{Hard}\\ \midrule[1pt]\smallskip
PC-DARTS & 0.956 & 0.945 & 0.881 & 0.957 & 0.947 & 0.882 & - & - & - \\ \hline\smallskip
ours+cat$\_$all & 0.957 & 0.947 & 0.883 & 0.957 & 0.948 & 0.884 & 0.957 & 0.948 & 0.885 \\ \hline\smallskip
ours+cat$\_$leaf & 0.957 & 0.947 & 0.885 & \textbf{0.958} & \textbf{0.949} & \textbf{0.887} & 0.958 & 0.948 & 0.887 \\ \bottomrule[2pt]

\end{tabular}
\begin{tablenotes}
\footnotesize
\item[*] ``cat$\_$all'' means all intermediate nodes are concatenated as the output, and ``cat$\_$leaf'' means only the leaf ones are concatenated.
\end{tablenotes}
\end{threeparttable}
}
\label{table:auto_cpm}
\vspace{-3mm}
\end{center}
\end{table}

Moreover, simulations are conducted to verify the effectiveness of our improved NAS approach for searching AutoFEM-CPM compared against PC-DARTS as shown in Table~\ref{table:auto_cpm}, where modules with $8$ intermediate nodes are only searched with our method due to the memory limitation. As we can see our improved method with 6 intermediate nodes achieves the greatest AP scores on the Easy, Medium and Hard subsets by concatenating the leaf nodes only.

\begin{table}[!t]
\small
\begin{center}
\caption{\small \textbf{Comparison of Average Precision (AP) of DRMC loss} in validation set of WIDER FACE. Multi-scale results ensemble is adopted during test-time.}
\resizebox{.75\columnwidth}{!}{
\smallskip\begin{tabular}{c|c|c|c}
\toprule[2pt]
Components & Easy & Medium & Hard  \\
\midrule[1pt]\smallskip
Baseline & 0.954 & 0.944 & 0.883  \\
\hline\smallskip
Baseline+Auxiliary loss  & 0.954 & 0.945 & 0.884 \\
\hline\smallskip
Baseline+Auxiliary loss+MC loss & 0.954 & 0.945 & 0.885 \\\hline\smallskip
Baseline+Auxiliary loss+DR loss & 0.955 & 0.946 & 0.883 \\ \hline\smallskip
Baseline+Auxiliary loss+DRMC loss &  \textbf{0.957} &  \textbf{0.947} &  \textbf{0.884} \\
Baseline+AutoFEM+Auxiliary loss+DRMC loss &  \textbf{0.961} &  \textbf{0.953} &  \textbf{0.888} \\
\bottomrule[2pt]
\end{tabular}
}
\vspace{-3mm}
\label{table:drmc}
\end{center}
\end{table}

\begin{table}[!t]
\small
\begin{center}
\caption{\small \textbf{Performance on WIDER FACE}. $\#Params$ and $\#FLOPS$ denote the number of parameters and multiply-adds. LAT denotes network inference latency with VGA resolution image.}
\resizebox{.95\columnwidth}{!}{
\smallskip\begin{tabular}{c|c|c|c|cc|cc|cc}
\toprule[2pt]
Model & Easy & Medium & Hard & \#Params & Ratio & \#FLOPS & Ratio & LAT(ms) & Ratio \\
\midrule[1pt]
\textbf{ASFD-D0}  & \textbf{0.901} & \textbf{0.875} & \textbf{0.744} & \textbf{0.62M} & \textbf{1x} & \textbf{0.73B} & \textbf{1x} & \textbf{8.9} & \textbf{1x} \\
EXTD(mobilenet)~\cite{yoo2019extd} & 0.851 & 0.823 & 0.672 & 0.68M & 1.1x & 10.62B & 14.5x & 34.4 & 3.9x\\\midrule[1pt]
\textbf{ASFD-D1}  & \textbf{0.933} & \textbf{0.917} & \textbf{0.820} & \textbf{3.90M} & \textbf{1x} & \textbf{4.27B} & \textbf{1x} & \textbf{9.2} & \textbf{1x} \\
SRN(Res50)~\cite{chi2019selective} & 0.930 & 0.873 & 0.713 & 80.18M & 20.6x & 189.69B & 44.4x &	55.1 & 6.0x\\\midrule[1pt]
\textbf{ASFD-D2}  & \textbf{0.951} & \textbf{0.937} & \textbf{0.836} & \textbf{13.56M} & \textbf{1x} & \textbf{20.48B} & \textbf{1x} & \textbf{12.4} & \textbf{1x} \\ 
Retinaface(Res50)~\cite{deng2019retinaface} & 0.957 & 0.943 & 0.828 & 26.03M & 1.9x & 33.41B & 2.4x & 29.3 & 2.4x \\\midrule[1pt]
\textbf{ASFD-D3}  & \textbf{0.953} & \textbf{0.943} & \textbf{0.848} & \textbf{26.56M} & \textbf{1x} & \textbf{46.32B} & \textbf{1x} & \textbf{23.1} & \textbf{1x}\\
PyramidBox(Res50)~\cite{tang2018pyramidbox} & 0.951 & 0.943 & 0.844 & 64.15M & 2.4x & 111.09B & 2.4x& 54.5 & 2.4x \\\midrule[1pt]
\textbf{ASFD-D4}  & \textbf{0.956} & \textbf{0.945} & \textbf{0.858} & \textbf{36.76M} & \textbf{1x} & \textbf{70.45B} & \textbf{1x} & \textbf{28.7} & \textbf{1x}\\
DSFD(Res152)~\cite{li2019dsfd} & 0.955 & 0.942 & 0.851 & 114.5M & 3.1x & 259.55B & 3.7x & 83.3 & 2.9x \\\midrule[1pt]
\textbf{ASFD-D5}  & \textbf{0.957} & \textbf{0.947} & \textbf{0.859} & \textbf{67.73M} & \textbf{1x} & \textbf{147.40B} & \textbf{1x} & \textbf{32.8} & \textbf{1x}\\\midrule[1pt]
\textbf{ASFD-D6}  & \textbf{0.958} & \textbf{0.947} & \textbf{0.860} & \textbf{86.10M} & \textbf{1x} & \textbf{183.11B} & \textbf{1x} & \textbf{37.7} & \textbf{1x}\\
\bottomrule[2pt]
\multicolumn{10}{c}{We omit ensemble and test-time multi-scale results, Latency are measured on the same machine.} \\
\end{tabular}
}
\vspace{-6mm}
\label{table:compound_scaling}
\end{center}
\end{table}

\vspace{-8pt}
\subsubsection{DRMC Loss.} We use DSFD~\cite{li2019dsfd} as the baseline to add Distance-based Regression and Margin-based Classification loss for comparison. As presented in Table~\ref{table:drmc}, the proposed DRMC loss together with the auxiliary one, that is, the loss operating on the output of the first shot brings the performance improvements of $0.3\%$, $0.3\%$ and $0.1\%$ on Easy, Medium and Hard subsets respectively for the DSFD baseline, and $0.3\%$, $0.4\%$ and $0.1\%$ for the AutoFEM-based DSFD.

\vspace{-8pt}
\subsubsection{Improved Model Scaling.} 
As discussed in Sec.~\ref{section:3.4}, an improved model scaling approach is proposed to make a trade-off between speed and accuracy by jointly scaling up depth and width of backbone, feature enhance module and head network of our ASFD. 
The comparisons of our ASFD D$0$-D$6$ with other methods are presented in Table~\ref{table:compound_scaling}, where our models achieve better efficiency than others, suggesting the superiority of AutoFEM searched by the improved NAS method and benefits of jointly scaling by balancing the dimensions of different architectures. 
In specific, our ASFD D$0$ and D$1$ can run at more than $100$ frame-per-second (FPS) on Nvidia P$40$ GPU with the lightweight backbone.
Even the model with the highest AP scores, \textit{e.g.} ASFD-D$6$, can run at $26$ FPS approximately, which is still $2.2$ times faster than DSFD with better performance.%

\begin{figure}[!tbp]
    \centering
    \subfigure[Val: Easy]{
        \includegraphics[height=0.23\linewidth]{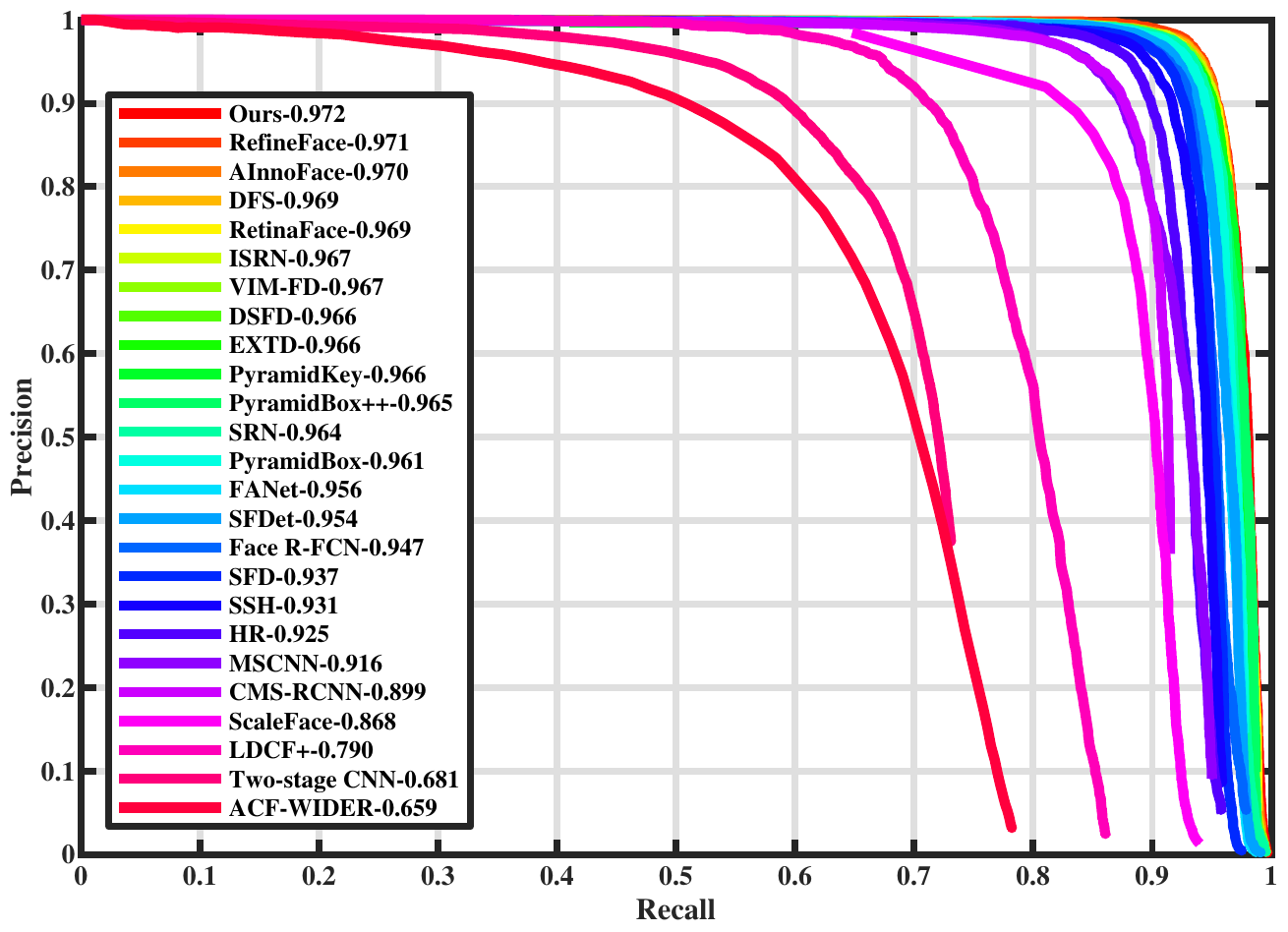}
    }
    \hspace{-6pt}
    \subfigure[Val: Medium]{
        \includegraphics[height=0.23\linewidth]{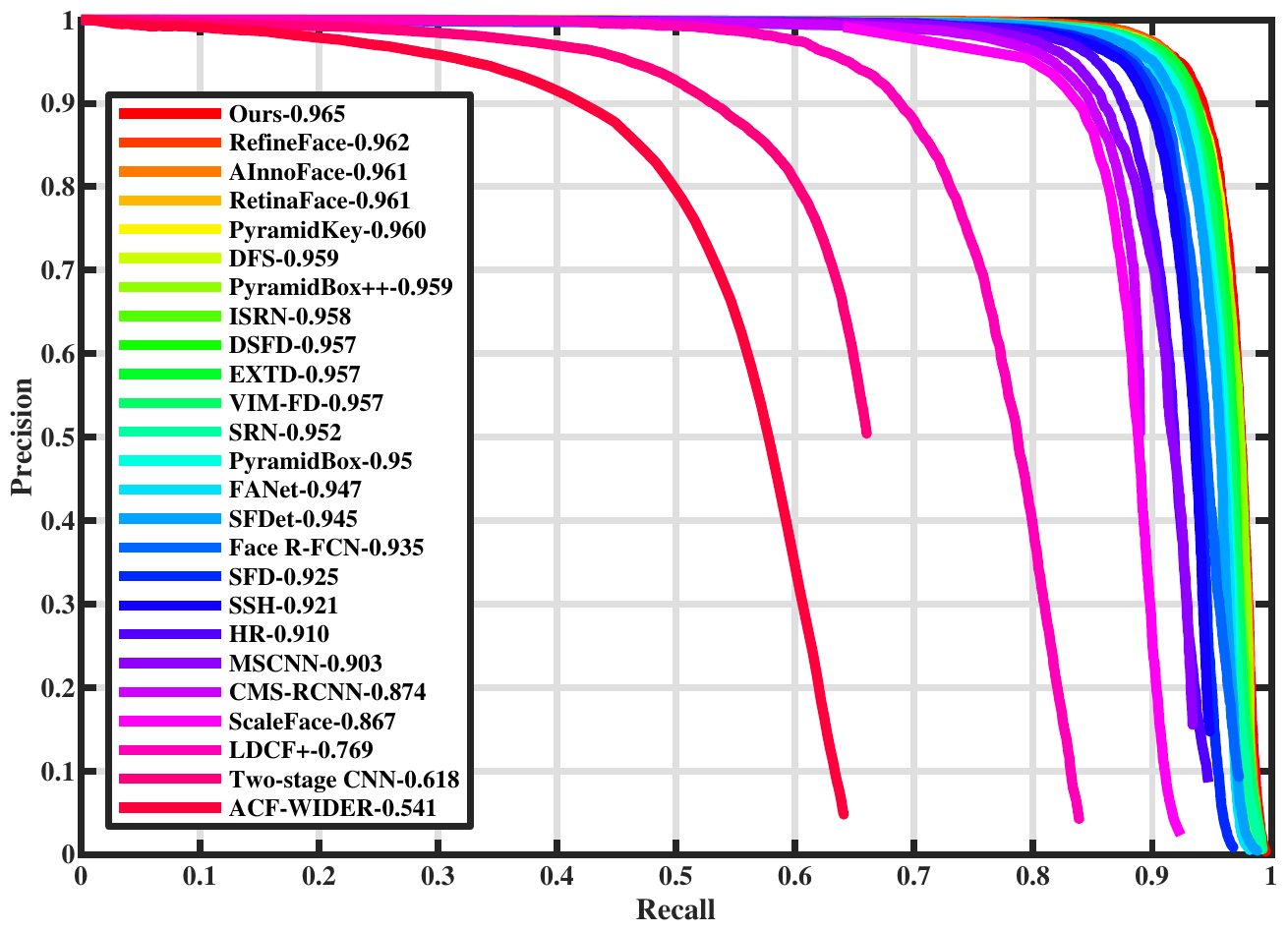}
    }
    \hspace{-6pt}
    \subfigure[Val: Hard]{
        \includegraphics[height=0.23\linewidth]{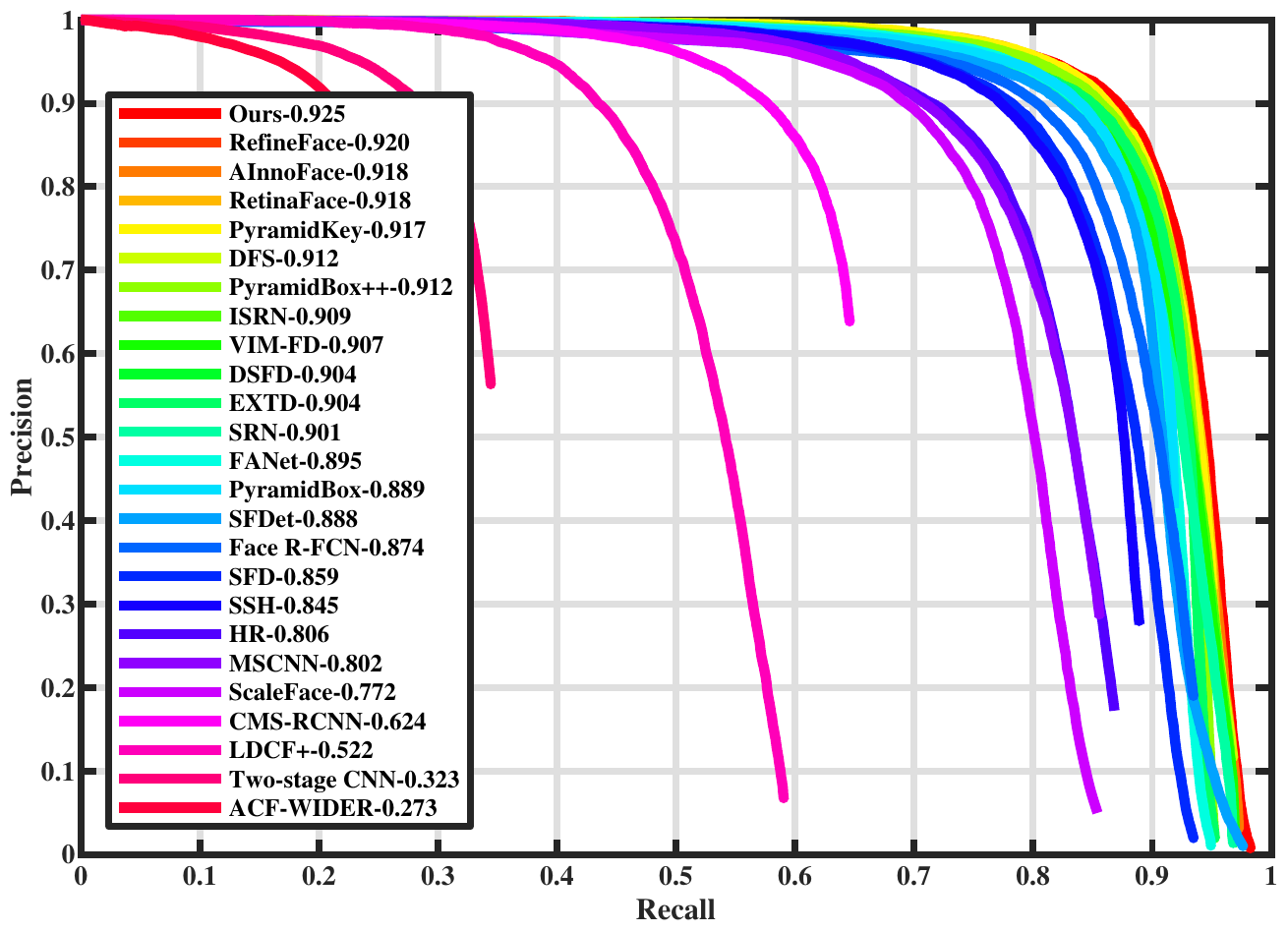}
    }
    \hspace{-6pt}
    \vfill
    \subfigure[Test: Easy]{
    	\includegraphics[height=0.23\linewidth]{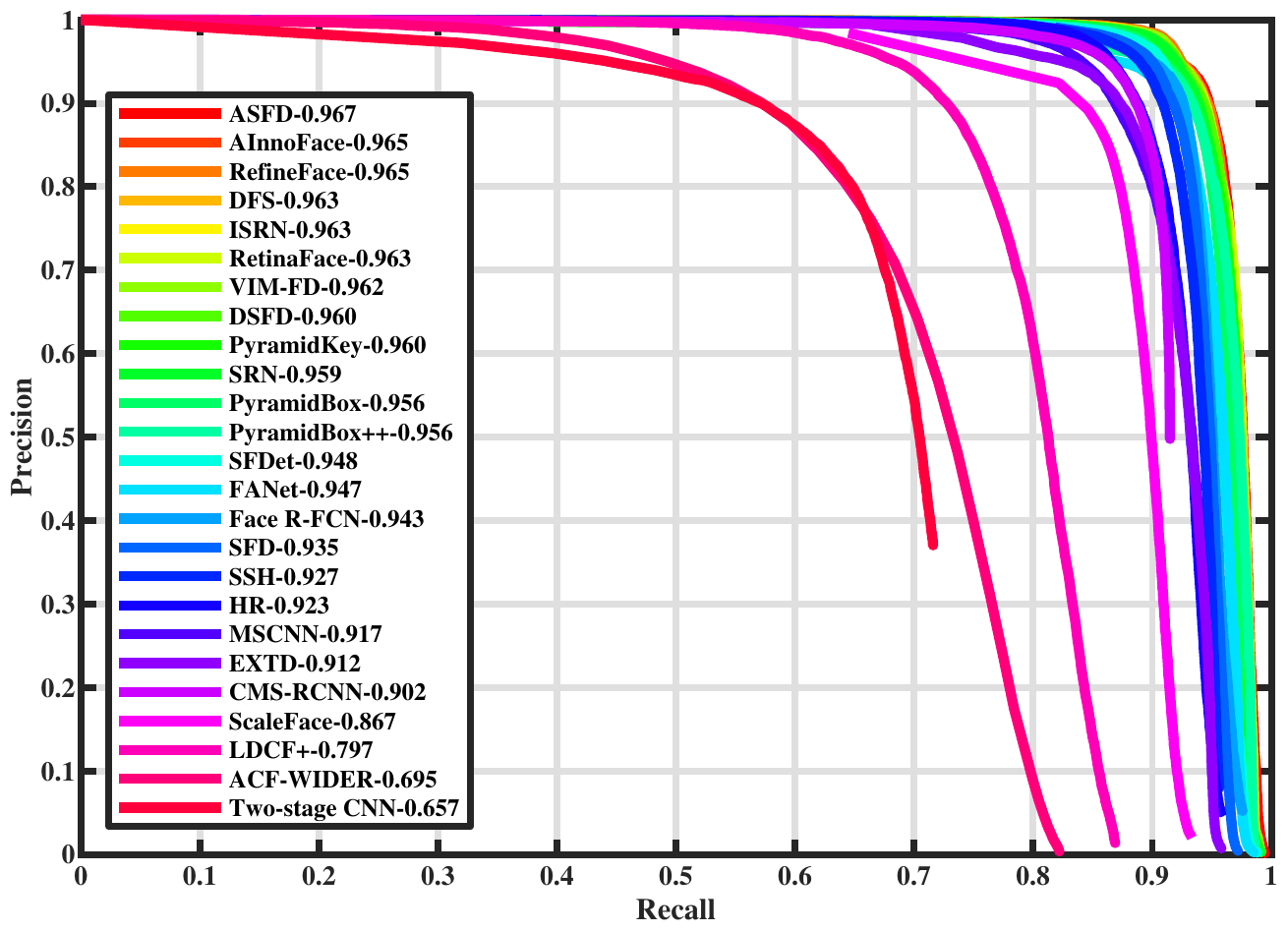}
    }
    \hspace{-6pt}
    \subfigure[Test: Medium]{
    	\includegraphics[height=0.23\linewidth]{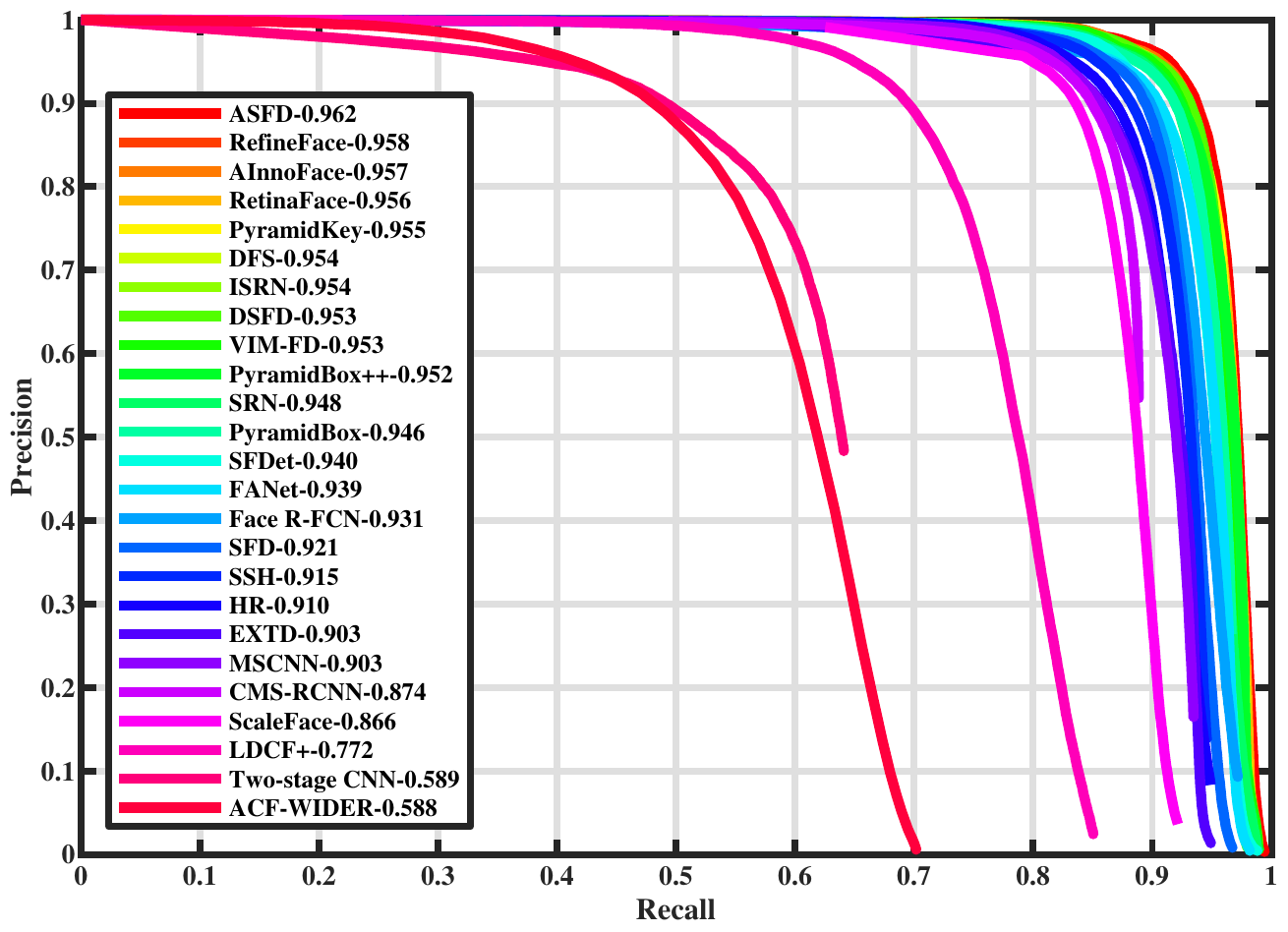}
    }
    \hspace{-6pt}
    \subfigure[Test: Hard]{
    	\includegraphics[height=0.23\linewidth]{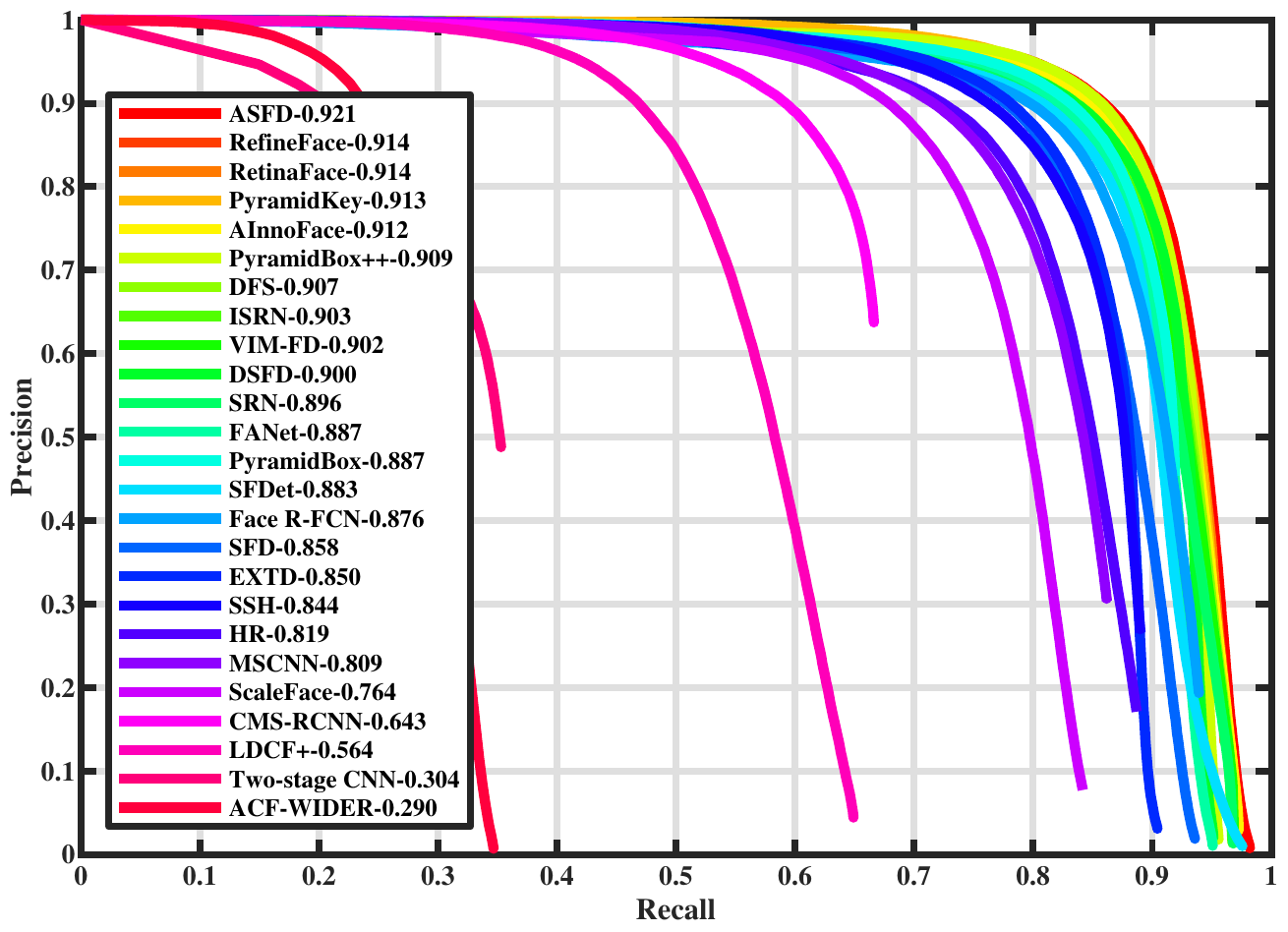}
    }
    \hspace{-6pt}
    \vfill
    \vspace{-3mm}
    \caption{\textbf{Precision-recall curves on WIDER FACE.}}
    \vspace{-4mm}
    \label{fig:WIDER_Face}
\end{figure}

\begin{figure}[!t]
    \centering
    \subfigure[Discontinuous ROC curves.]{
        \includegraphics[width=0.46\linewidth]{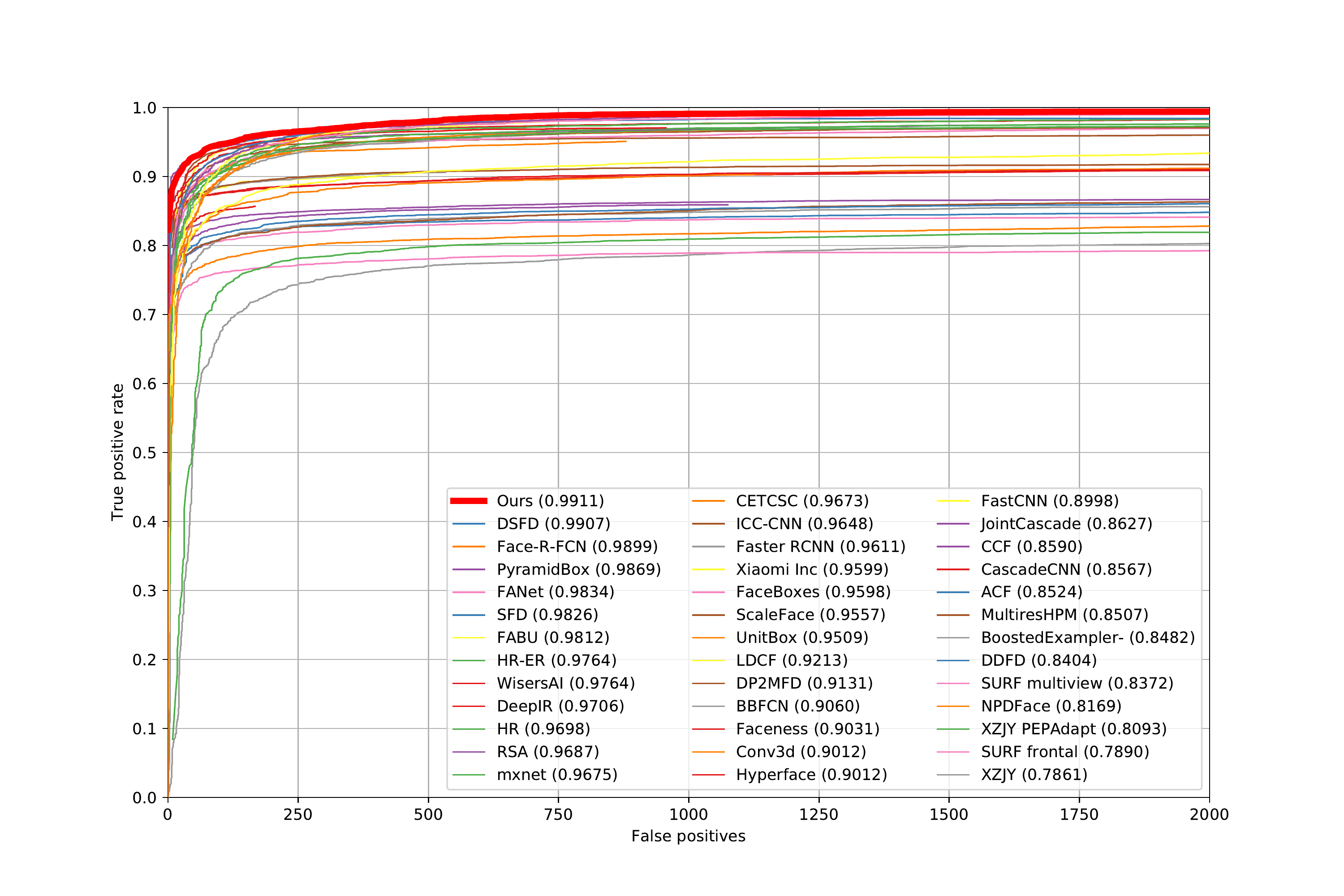}
    }
    \subfigure[Continuous ROC curves.]{
        \includegraphics[width=0.46\linewidth]{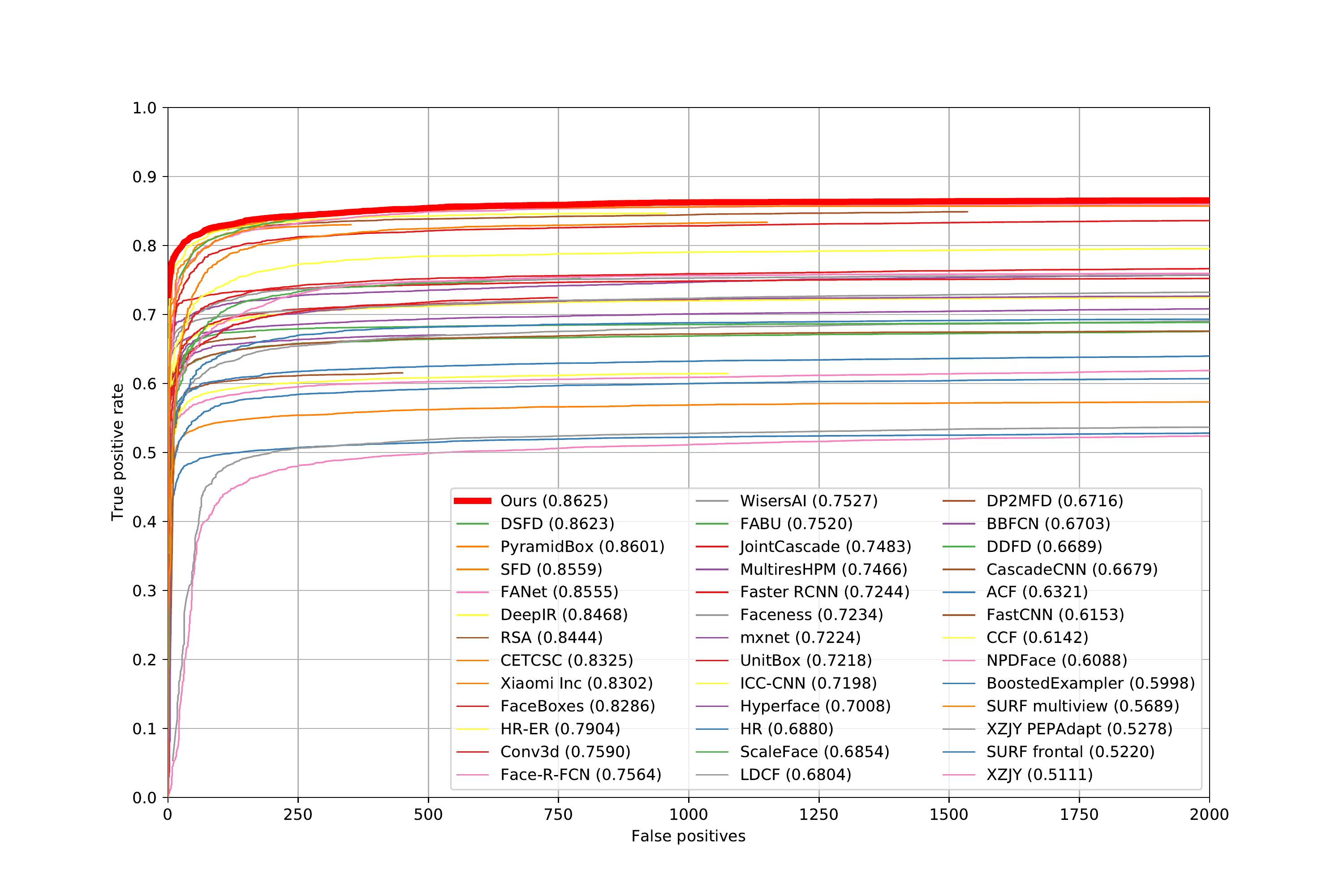}
    }
    \vspace{-3mm}
    \caption{\small \textbf{ROC curves on the FDDB dataset}.}
    \label{fig:FDDB}
\end{figure}

\subsection{Comparisons with State-of-the-Art Methods}
\begin{figure}[!t]
    \centering
    \includegraphics[width=0.325\linewidth]{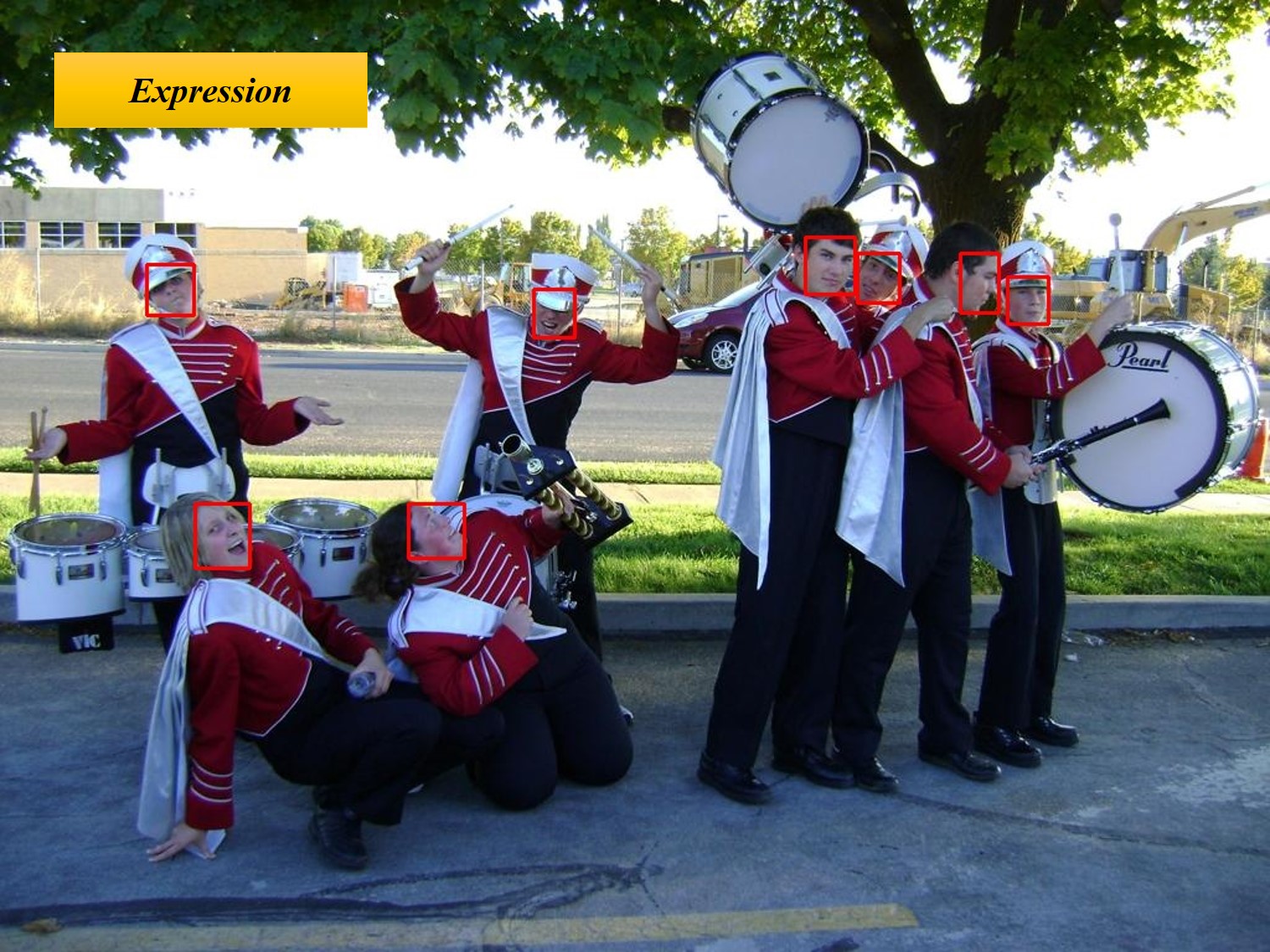}
    \includegraphics[width=0.325\linewidth]{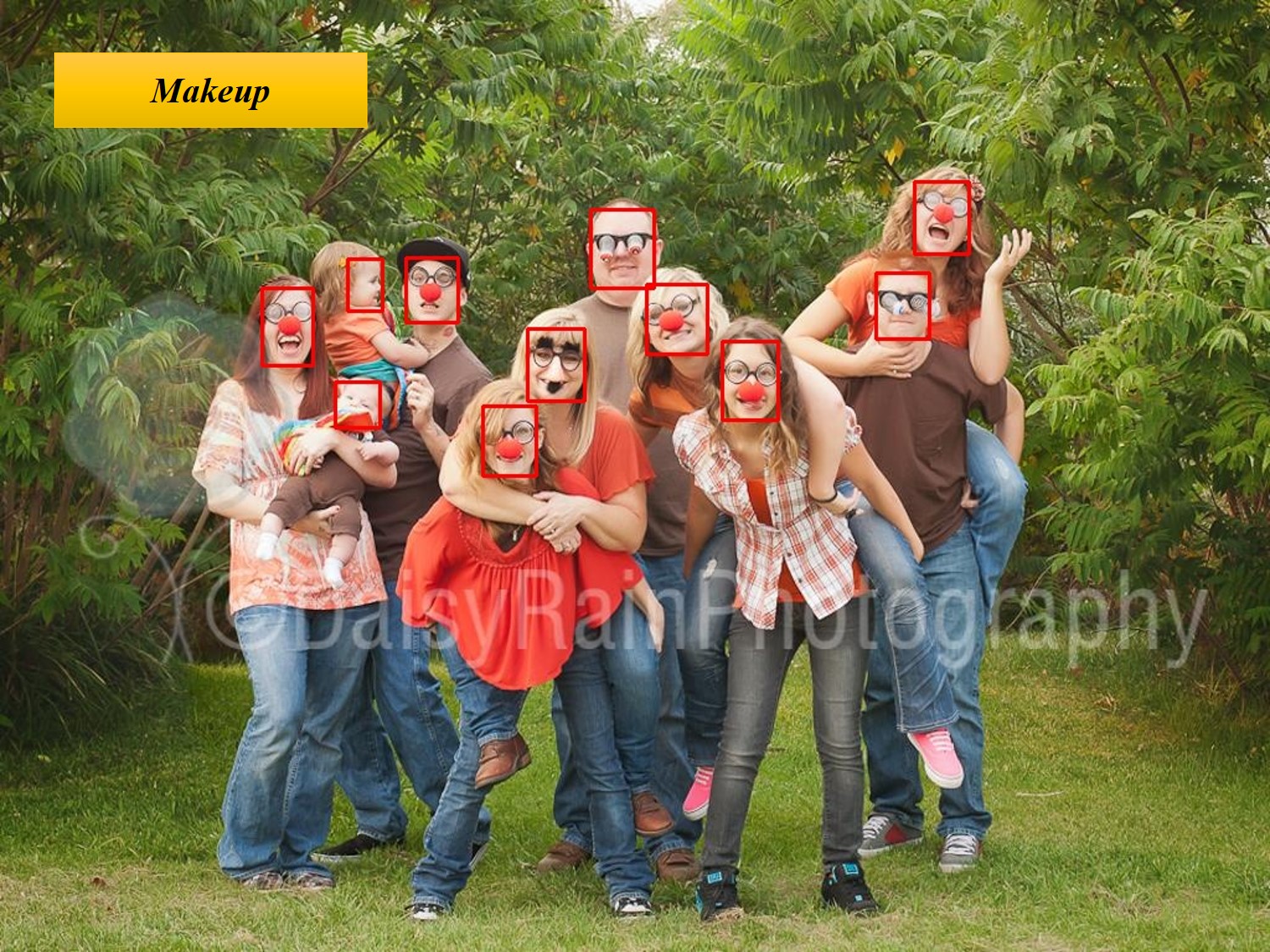}
    \includegraphics[width=0.325\linewidth]{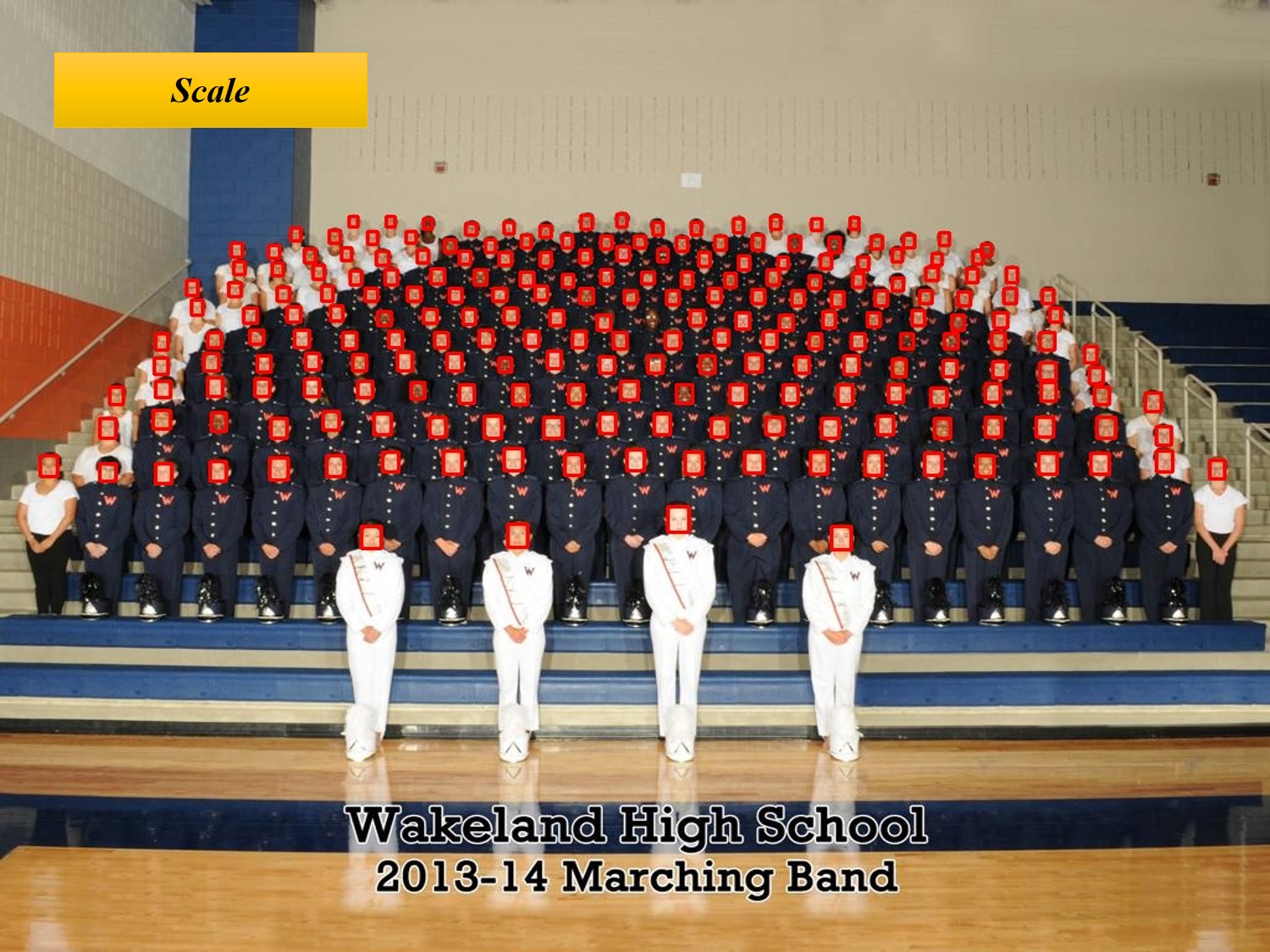}
    \vfill
    \includegraphics[width=0.325\linewidth]{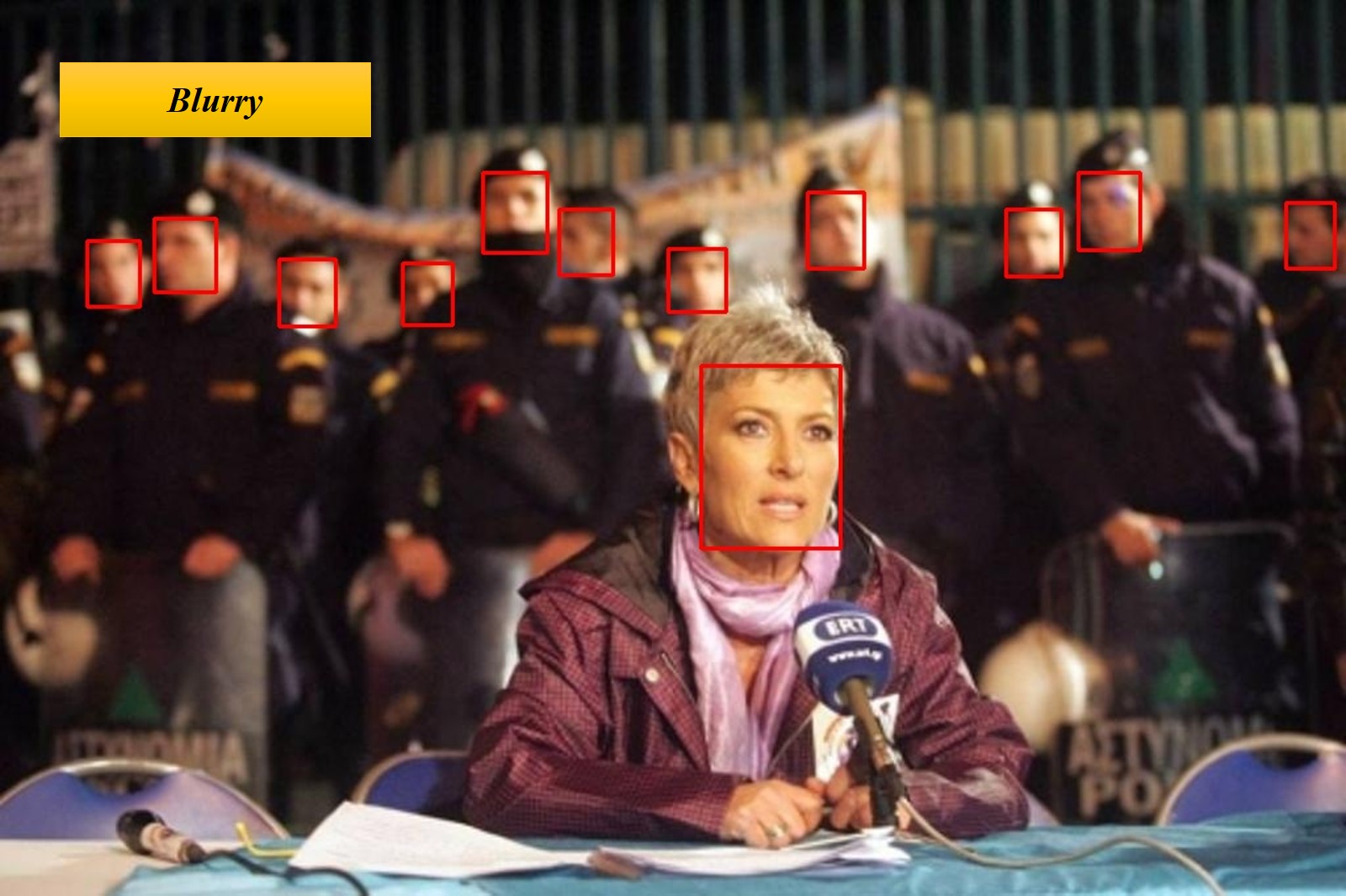}
    \includegraphics[width=0.325\linewidth]{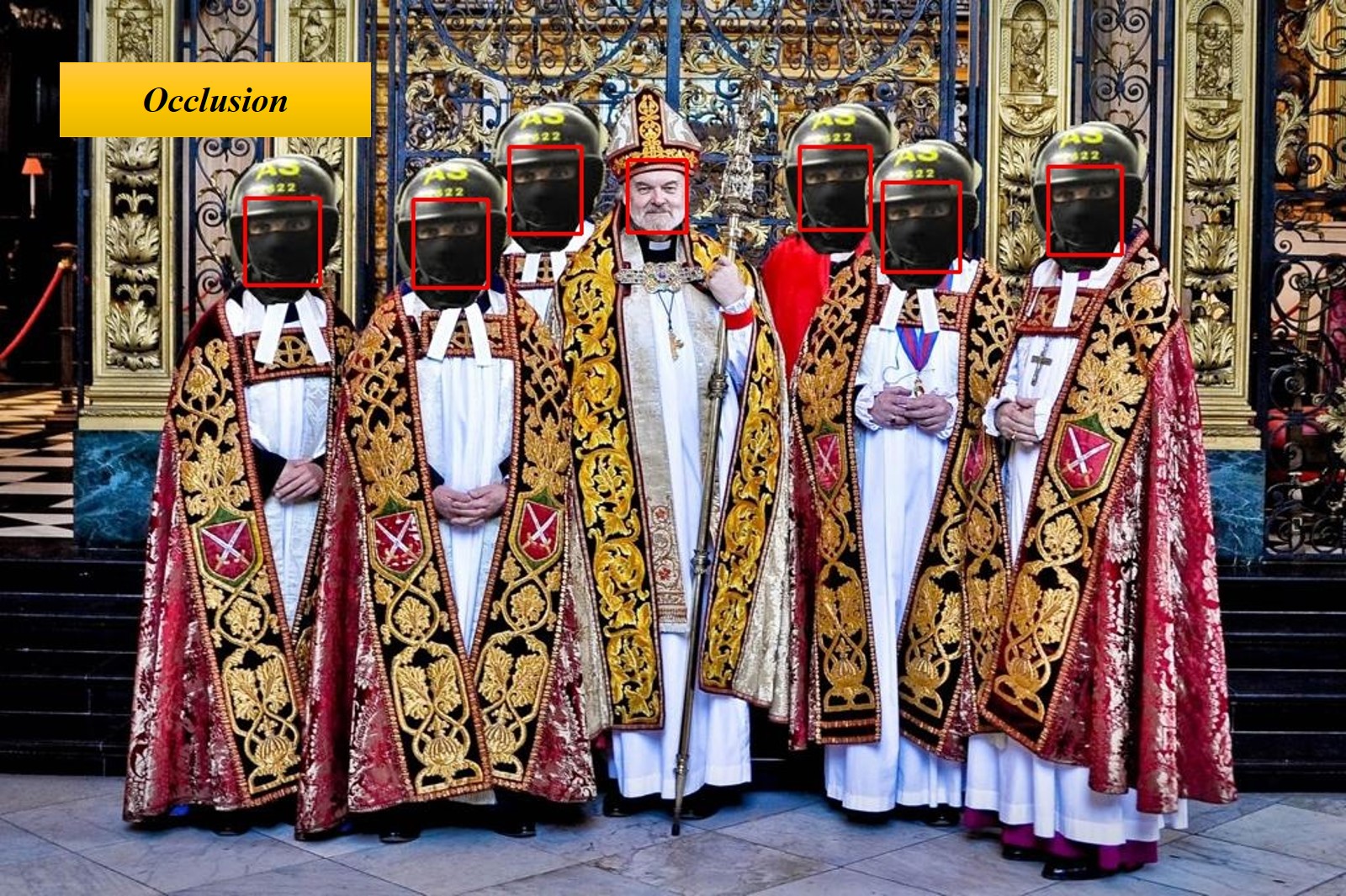}
    \includegraphics[width=0.325\linewidth]{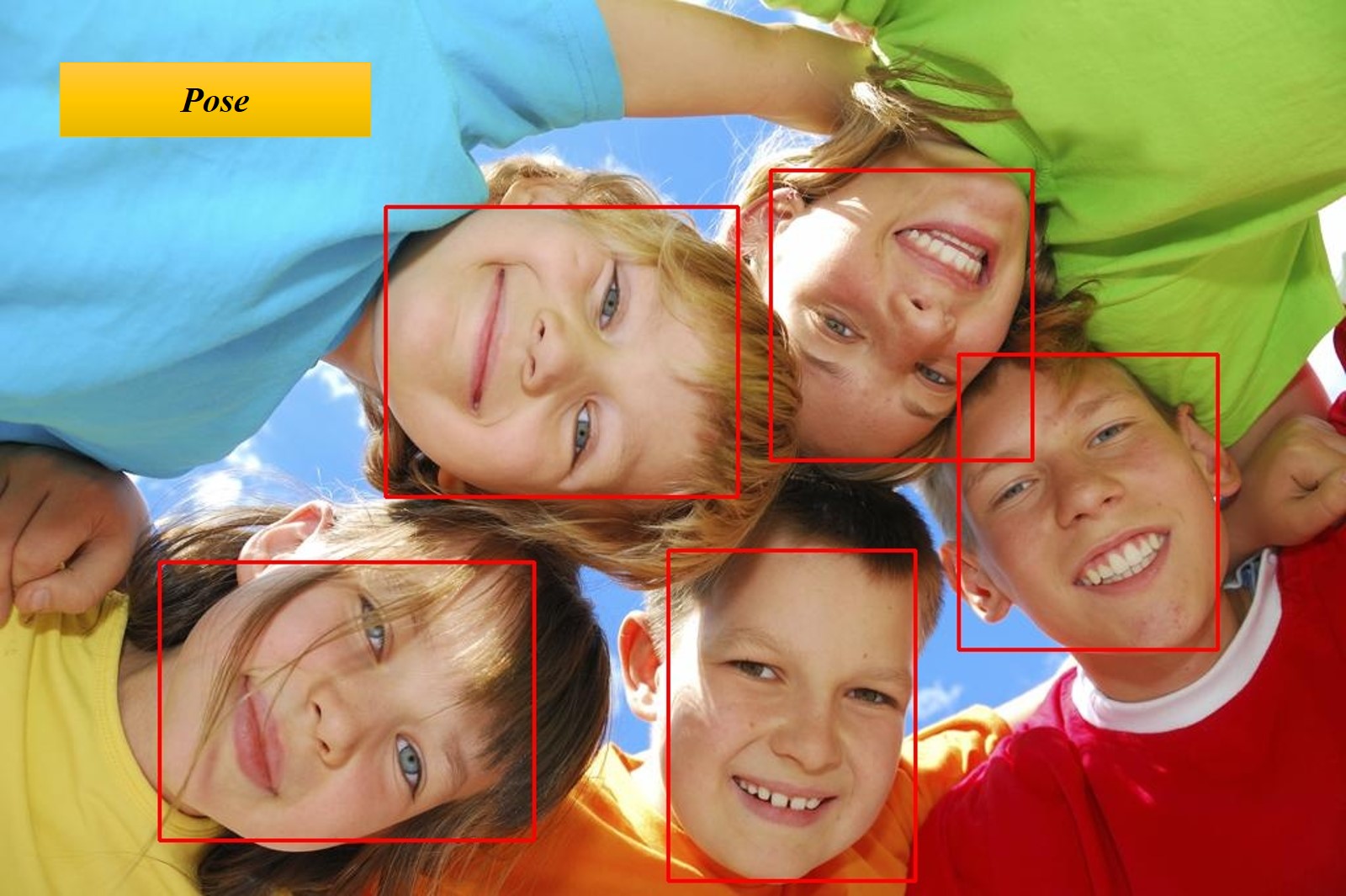}
    \vfill
    \includegraphics[width=0.325\linewidth]{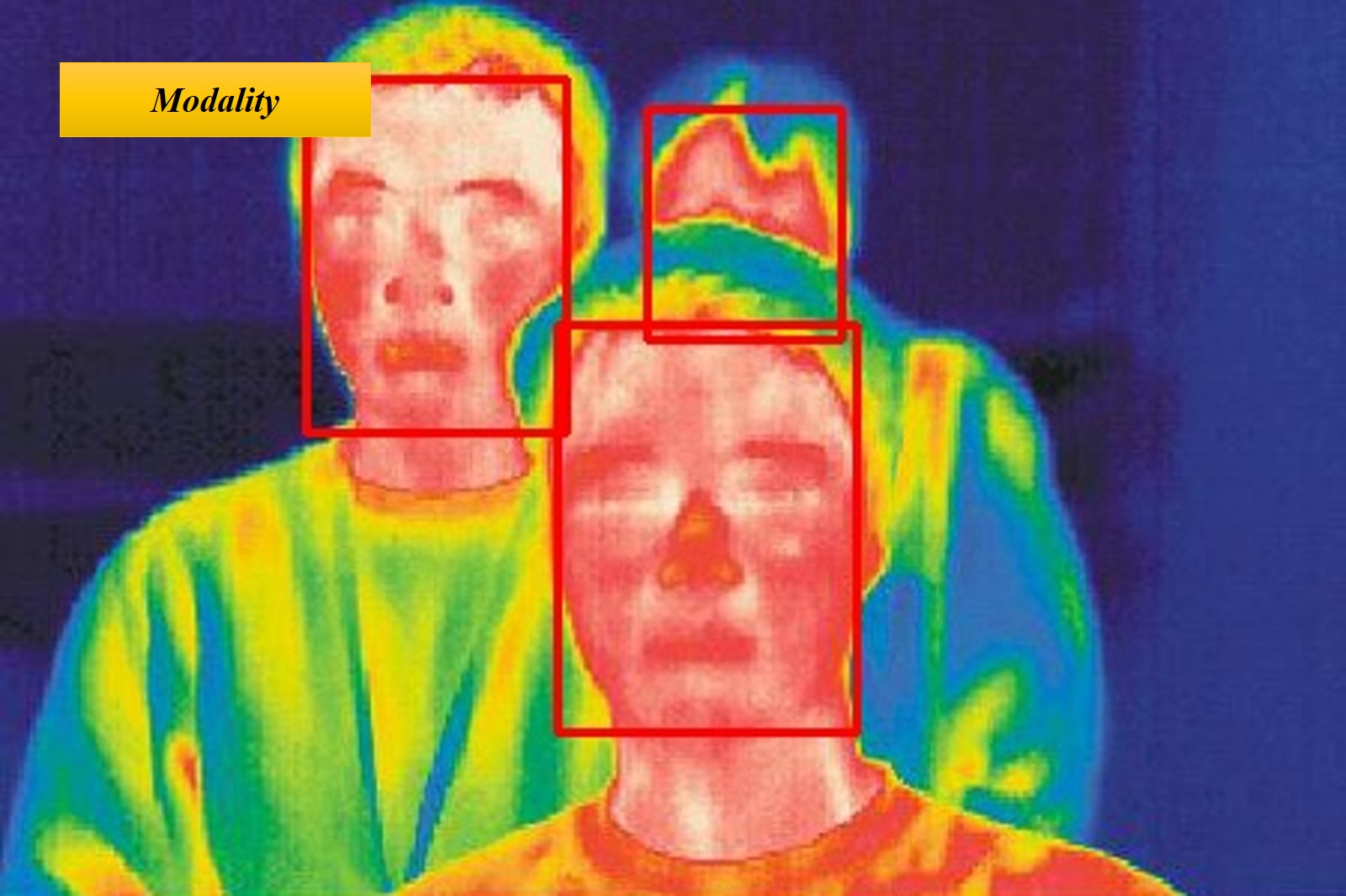}
    \includegraphics[width=0.325\linewidth]{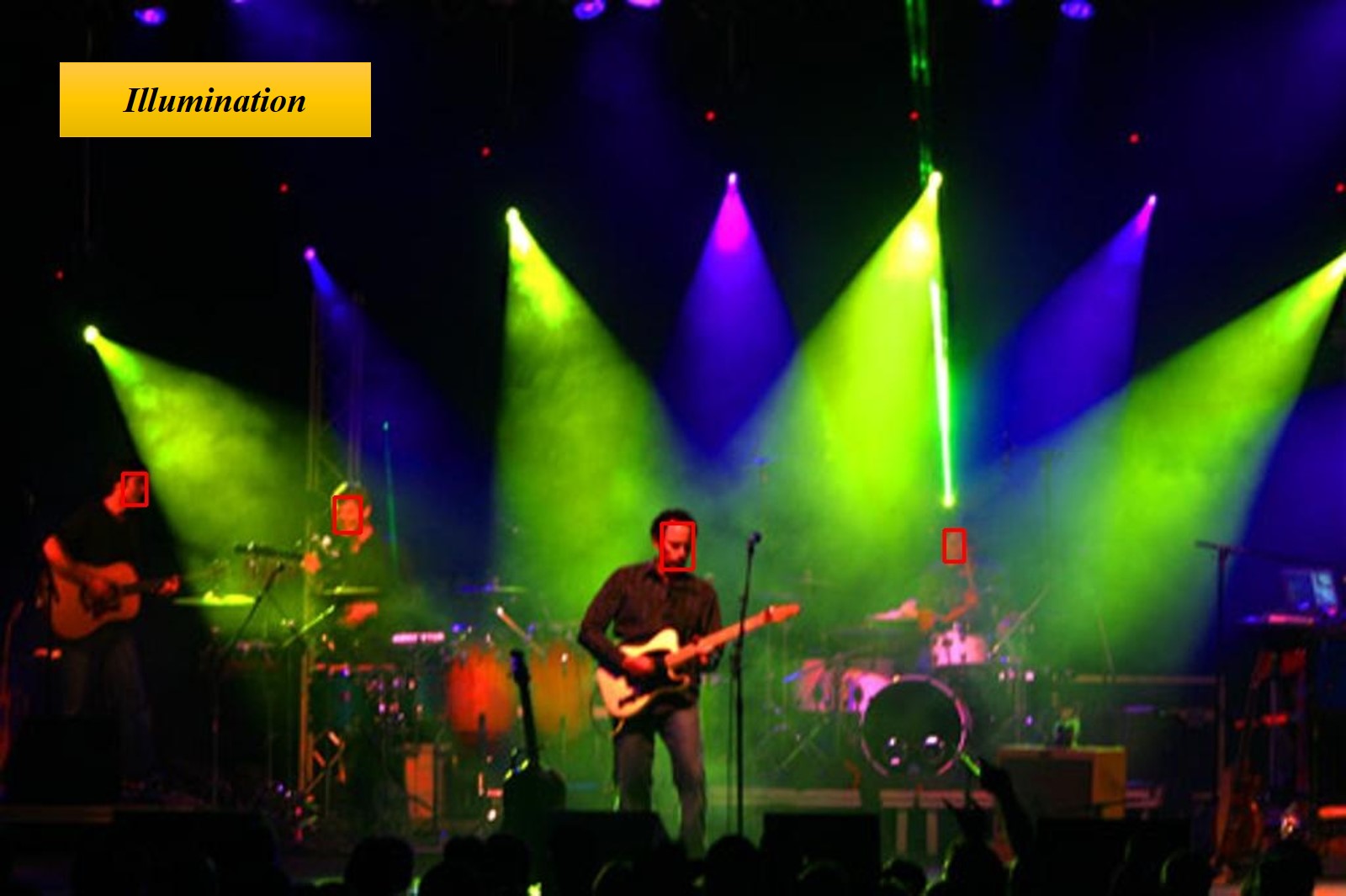}
    \includegraphics[width=0.325\linewidth]{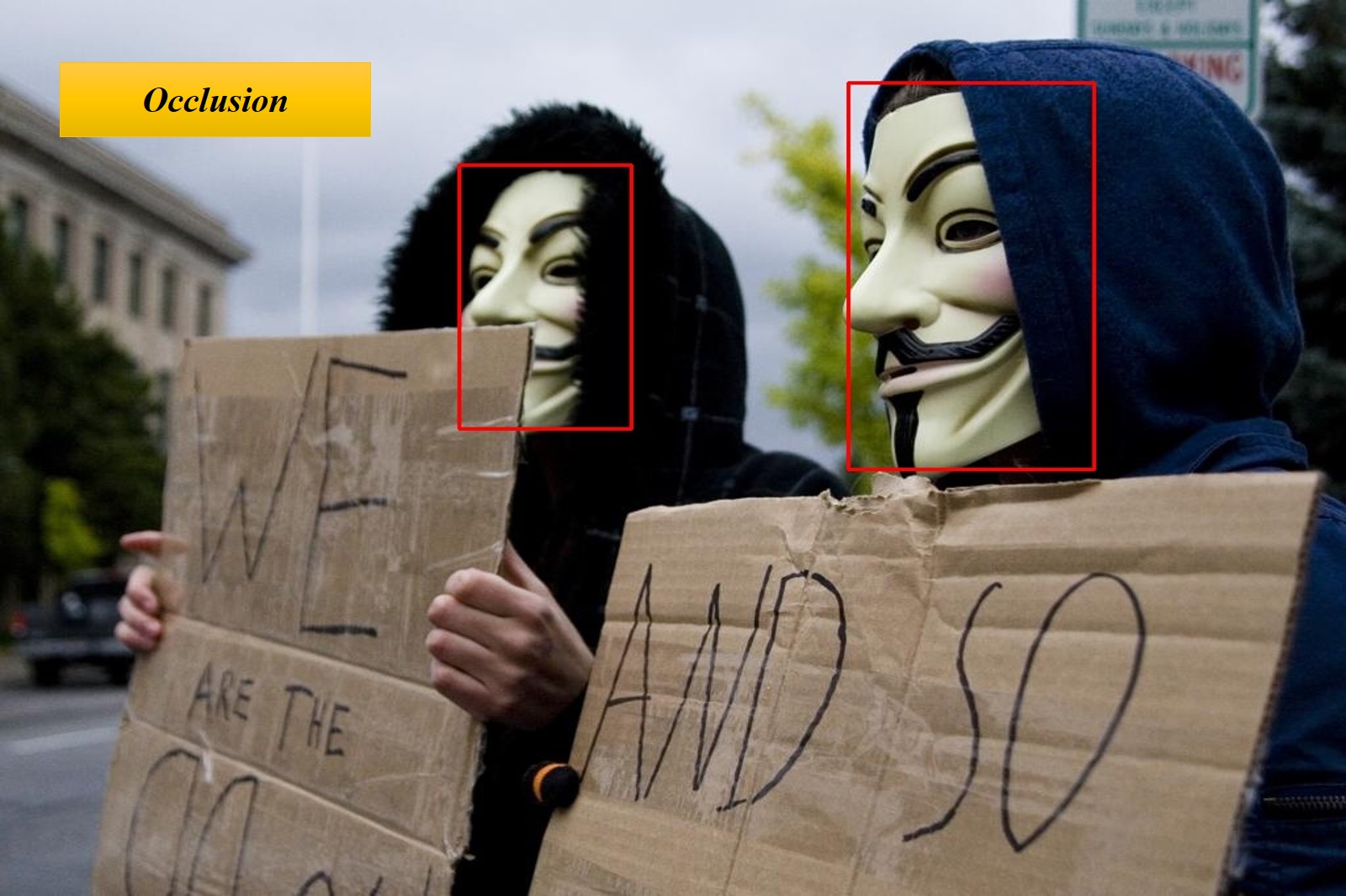}
    \vfill
    \vspace{-2.5mm}
    \caption{\small \textbf{Illustration of our ASFD to various large variations}. Red bounding boxes indicate the detection confidence is above $0.8$.}
    \vspace{-3mm}
    \label{fig:visual_demo}
\end{figure}

Finally, we evaluate our ASFD on two popular benchmarks, WIDER FACE~\cite{yang2016wider} and FDDB ~\cite{jain2010fddb} using ASFD-D$6$. 
Our model is trained \textit{only} on the training set of WIDER FACE and evaluated on both benchmarks without any fine-tuning. 
We also follow the setting in~\cite{li2019dsfd} to build image pyramids for multi-scale testing for better performance. 
Our ASFD-D$6$ obtains the highest AP scores of $97.2\%$, $96.5\%$ and $92.5\%$ on the Easy, Medium and Hard subsets of WIDER FACE validation, as well as $96.7\%$, $96.2\%$ and $92.1\%$ on test, as shown in Fig.~\ref{fig:WIDER_Face}, setting a new state-of-the-art face detector, meanwhile, the ASFD-D$6$ is faster than Refineface (37.7 vs 56.6 ms) even it is our best competitor in performance~\cite{zhang2019refineface}.
The state-of-the-art performance is also achieved on FDDB, \textit{i.e.}, $99.11\%$ and $86.25\%$ true positive rates on discontinuous and continuous curves when the number of false positives is $1000$, as shown in Fig.~\ref{fig:FDDB}. 
More examples of our ASFD on handling face with various variations are shown in Fig.~\ref{fig:visual_demo} to demonstrate its effectiveness.

\section{Conclusions}
In this work, a novel Automatic and Scalable Face Detector (ASFD) is proposed with significantly better accuracy and efficiency, in which we adopt a differential architecture search to discover feature enhance modules for efficient multi-scale feature fusion and context enhancement. Besides, the Distance-based Regression and Margin-based classification (DRMC) losses are introduced to effectively generate accurate bounding boxes and highly discriminative deep features. We also adopt improved model scaling methods to develop a family of ASFD by scaling up and down the backbone, feature module, and head network. Comprehensive experiments conducted on popular benchmarks FDDB and WIDER FACE to demonstrate the efficiency and accuracy of our proposed ASFD compared with state-of-the-art methods.

\clearpage
%
%
\bibliographystyle{splncs04}
\bibliography{egbib}
\end{document}